\documentclass[11pt]{article}

\usepackage[preprint]{acl}

\usepackage{times}
\usepackage{latexsym}

\usepackage[T1]{fontenc}

\usepackage[utf8]{inputenc}

\usepackage{microtype}

\usepackage{inconsolata}

\usepackage{float}
\usepackage{graphicx}
\usepackage{amsmath}
\usepackage{booktabs}
\usepackage{tabularx}
\usepackage{float}
\usepackage{multicol, multirow}
\usepackage{tcolorbox}
\usepackage{enumitem}
\usepackage{textcomp}

\usepackage[ruled,vlined]{algorithm2e}

\definecolor{MySoftGreen}{RGB}{34,139,34}
\definecolor{MyPurple}{RGB}{104, 52, 154}

\newcommand{\n}{{\sf ANCHOR}}

\newcommand{\s}{\sf SAFE}

\newcolumntype{Y}{>{\centering\arraybackslash}X}

\title{\n: LLM-driven Subject Conditioning for Text-to-Image Synthesis}

\author{Aashish Anantha Ramakrishnan\textsuperscript{1}~~~~
Sharon X. Huang\textsuperscript{2}~~~~
\textbf{Dongwon Lee\textsuperscript{2}}\\[0.5em]
  \textsuperscript{1}Optum AI~~~~
  \textsuperscript{2}The Pennsylvania State University\\
\textsuperscript{1}\texttt{aashish.anantha.ramakrishnan@optum.com}~~~\\
\textsuperscript{2}\texttt{\{suh972, dul13\}@psu.edu}\\
}

\begin{document}
\maketitle
\begin{abstract}
Text-to-image (T2I) models have achieved remarkable progress in high-quality image synthesis, yet most benchmarks rely on simple, self-contained prompts, failing to capture the complexity of real-world captions. Human-written captions often involve multiple interacting subjects, rich contextual references, and abstractive phrasing, conditions under which current image-text encoders like CLIP struggle. To systematically study these deficiencies, we introduce {\n}, a large-scale dataset of 70K+ abstractive captions sourced from five major news media organizations. Analysis with {\n} reveals persistent failures in multi-subject understanding, context reasoning, and nuanced grounding. Motivated by these challenges, we propose \emph{Subject-Aware Fine-tuning} ({\s}), which uses Large Language Models (LLMs) to extract key subjects and enhance their representation at the embedding-level. Experiments with contemporary models show that {\s} significantly improves image-caption consistency and human preference alignment, serving as a practical and scalable solution. The Dataset and code are available at: \url{https://github.com/aashish2000/ANCHOR}.
\end{abstract}

\begin{figure}[!ht]
      \centering
      \includegraphics[width=\linewidth]{"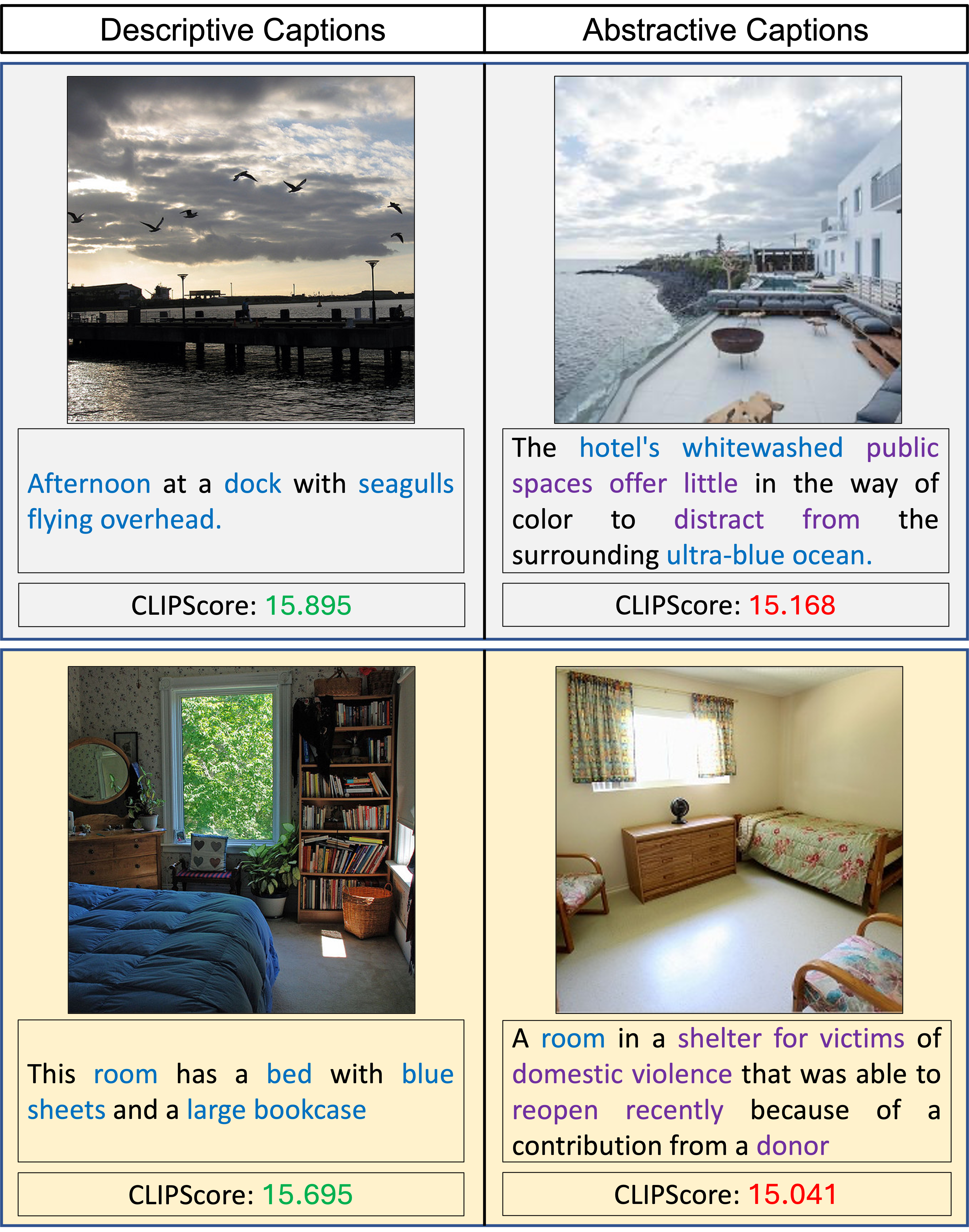"}
      \caption{Example of descriptive prompts from the COCO Captions dataset \cite{Chen2015-qj} (Left) and abstractive captions from the {\n} (Right). Words highlighted in \textcolor{blue}{Blue} directly translate to semantic objects while words highlighted in \textcolor{MyPurple}{Purple} signify contextual cues and syntactic variations, influencing the image indirectly. We also highlight abstractive captions differ at an embedding-level compared to descriptive prompts when paired with semantically similar images with lower CLIPScores.}
      \label{fig:abstractive_captions}
  \end{figure}

\section{Introduction}
\label{sec:intro}

Generative AI capabilities have grown rapidly, where large-scale pretraining has enabled powerful text-to-image (T2I) generation systems \cite{Wang2023-ve}. Yet, a fundamental challenge remains in aligning these systems with how language is naturally used in high-information settings. Most state-of-the-art T2I models are trained and evaluated on literal, object-centric prompts designed to directly reflect the visible contents of an image \cite{Sharma2018-tr, Chen2015-qj}. However, in many real-world contexts such as journalism, education, and social media, captions are often \textbf{abstractive}--i.e., they embed background knowledge, reference multiple entities, and convey discourse-level intent \cite{Grice1975-mh, Alikhani2019-kn, Vedantam2017-fa}.

In this paper, we position abstractive caption understanding as a critical yet underexplored challenge for T2I models and image-text encoders more broadly \cite{Liao2024-un}. While much progress has been made in aligning literal descriptions with visual content, real-world usage of Vision Language Models (VLM) require them to interpret language that leverages context information with implicit and narrative framings \cite{Song2021-ln}. With CLIP \cite{Radford2021-ro} being one of the most popular image-text encoders utilized for various VLMs \cite{Schlarmann2024-mm}, our goal is to analyze how CLIP-based T2I pipelines handle this form of complexity and provide actionable insights for improvement. Existing studies isolate specific reasoning tasks such as multi-object scenarios \cite{Abbasi2025-wa} and spurious image-text correlations \cite{Wang2024-az} but aren't representative of real-world data domains. This is important as image-caption pairs may utilize multiple reasoning sub-tasks for intent comprehension.

To support this analysis, we construct \textbf{{\n}}, a large-scale dataset of over 70K abstractive image-caption pairs curated from news media. These captions reflect how humans naturally write about images in context: not just describing what is seen, but linking it to entities, events, and narrative discourse as depicted in Figure \ref{fig:abstractive_captions}. The dataset is annotated to highlight four core challenges in real-world captioning: (1) \textit{Semantic Objects}, (2) \textit{Named Entities (NE)}, (3) \textit{Contextual Cues}, and (4) \textit{Syntactic Variations}. 

Our empirical analysis reveals that current CLIP-based T2I models struggle with multi-subject understanding, context disambiguation and entity resolution in these settings. To better diagnose and address these issues, we propose \textit{Subject-Aware FinE-tuning (\s)}, a lightweight, modular strategy that augments T2I conditioning using \textbf{LLM-extracted subject guidance}. Rather than retraining encoders or introducing larger architectures, {\s} operates as a plug-and-play module that helps models differentiate between core semantic objects from contextual cues along with improved generalization on diverse caption styles. Thus, {\s} helps improve alignment without sacrificing scalability. Our primary goal is to use this framework as a lens to understand the specific failure modes of current systems when confronted with abstractive, multi-subject language. Our contributions are as follows:

\begin{itemize}[leftmargin=*]
    \item We introduce {\n}, the first large-scale dataset of abstractive, real-world captions for probing the limitations of current T2I models and image-text encoders.
    \item We develop {\s}, a subject-aware fine-tuning method that highlights the importance of disentangling semantic objects from contextual modifiers in abstractive captions.
    \item We provide a detailed evaluation of CLIP-based T2I models on {\n}, showing where and why they fall short—and how {\s} can serve as a targeted, interpretable remedy.
\end{itemize}
\begin{figure*}
      \centering
      \includegraphics[width=0.75\textwidth]{"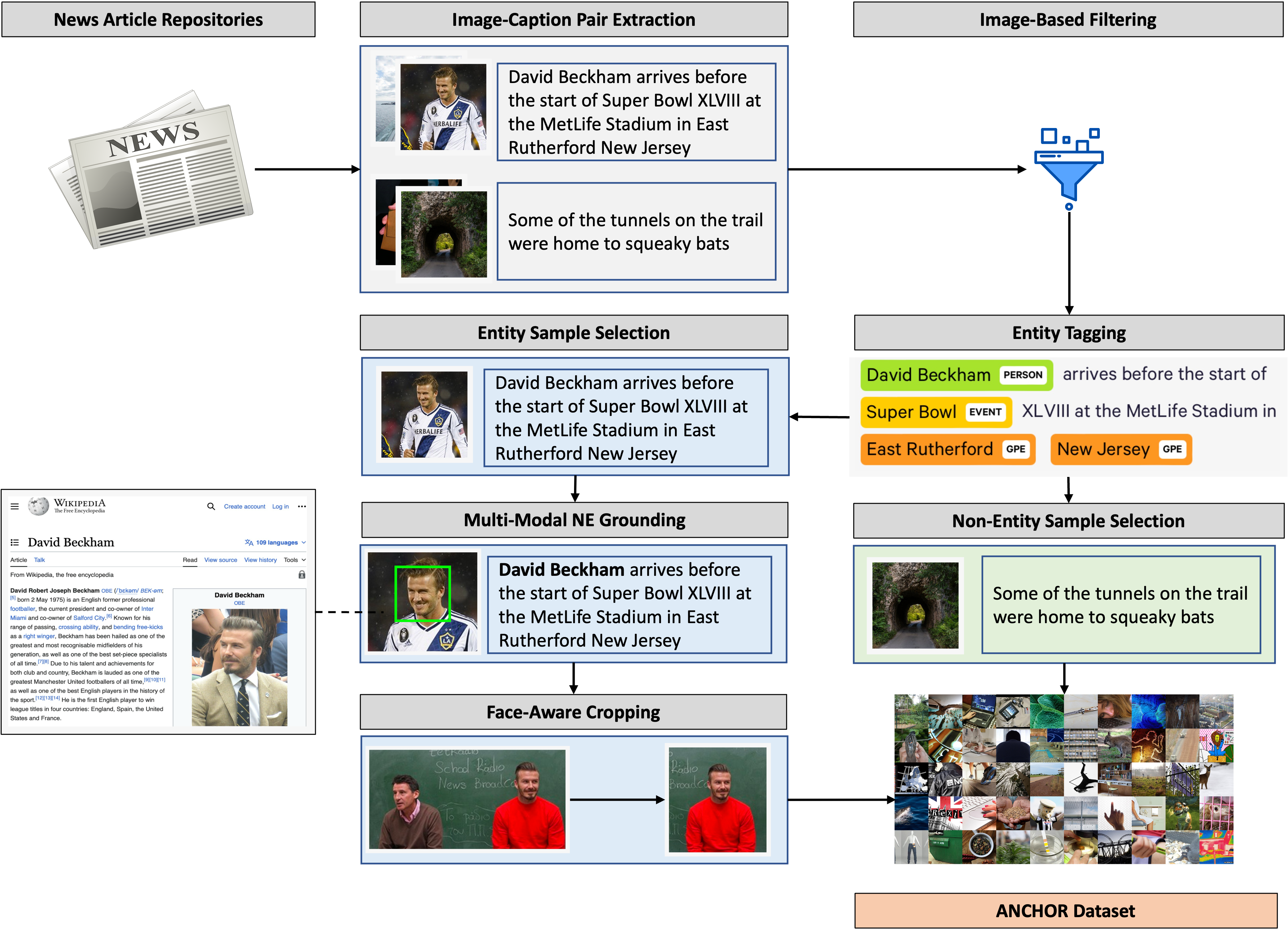"}
      \caption{Overview of our dataset's pre-processing and filtering steps for both {\n} Non-Entity and {\n} Entity Subsets.}
      \label{fig:dataset_overview}
  \end{figure*}

\section{Related Work}

\paragraph*{Text-to-Image Synthesis}
There have been significant improvements in the field of T2I generation since the launch of Generative Adversarial Networks (GAN) \cite{Goodfellow2014-zz, Zhang2017-kf, Xu2018-rp, Zhu2019-xs}. As Transformer-based architectures \cite{Vaswani2017-ju} and model pre-training was successful on Natural Language Processing tasks, multi-modal encoders such as CLIP \cite{Radford2021-ro} significantly improved the quality of multi-modal embeddings and provided better input conditioning \cite{Crowson2022-cp, Zhou2022-kn}. Diffusion models \cite{Sohl-Dickstein2015-je} provided a breakthrough in training higher resolution models with greater expressivity by modeling generation as a reverse-Markov chain process \cite{Nichol2022-hg}, \cite{Ramesh2021-ry, Ding2021-ry}. With the success of language model-based text-only encoders, Large Vision models adopt Large Language Model (LLM) based encoders for T2I generation, leveraging their text comprehension capabilities \cite{Saharia2022-ui}. Flow matching-based models have also shown promise for data domain-agnostic generation tasks \cite{Wang2025-xm}. 

\paragraph*{Multi-modal Reasoning}
Broadly, the types of reasoning tasks that models are commonly evaluated on are: Visual Reasoning, Context-based Reasoning, Factual Reasoning and Inter-modal Reasoning \cite{Li2024-bt, Anantha-Ramakrishnan2025-vl}. With Visual Reasoning, MLLMs are assessed based on their ability to incorporate specific visual cues in a structured manner for tasks such as spatial and object relationship understanding \cite{Thrush2022-yf, Li2024-sa, Kamath2023-ls, Zhang2025-uy}. The retrieval and interpretation of information from various sources related to specialized topics constitutes as Factual Reasoning \cite{Lu2022-aw, Wang2024-hy, Johnson2017-py}. On the other hand, Context-based Reasoning explicitly measures how well models make use of the provided in-context samples for tasks involving logical and compositional understanding \cite{Schwenk2022-gh, Zeng2024-sp, Zong2024-xa}. Finally, Inter-modal reasoning involves interpreting causal linkages between modalities based on both semantic features, pragmatic cues and commonsense knowledge \cite{Alikhani2020-nr, Xu2022-ie, Anantha-Ramakrishnan2025-zv,  Anantha-Ramakrishnan2025-je}. Our proposed {\n} dataset evaluates both the context-based and inter-modal reasoning capabilities of T2I models using abstractive captions. 

\paragraph*{Datasets}
Initial benchmarks for T2I models focused on utilizing open domain images with short, descriptive text for evaluating their generation performance \cite{Chen2015-qj}. In order to scale up diversity of evaluation prompts, image-text pairs crawled from online articles became more commonplace for facilitating both training and benchmarking these models \cite{Sharma2018-tr, Changpinyo2021-qp, Schuhmann2021-mp, Schuhmann2022-er, Schuhmann-C-Kopf-A-Vencu-R-Coombes-T-and-Beaumont-R2022-jy}. With T2I models being pre-trained on larger and larger corpora, there has been a shift towards evaluation-only benchmarks with prompts to judge specific attributes of a generator's performance. PartiPrompts \cite{Yu2022-yv}, DrawBench \cite{Saharia2022-ui} and UniBench \cite{Li2022-tf} provide diverse text prompts sorted based on style and difficulty. DiffusionDB \cite{Wang2023-ff} is a large-scale collection of prompt-tuned caption-image pairs commonly used for sourcing captions for T2I evaluation.  All these benchmarks focus on captions that provide accurate descriptions of physical objects within images. We aim to include captions containing situational context information and complex sentence structures as a part of {\n}.

\section{{\n}: Dataset Construction}
\label{sec:news_img_gen_overview}

The {\n} (Abstractive News Captions with High-level cOntext Representation) dataset is a large-scale image-caption pair dataset extracted from news articles as shown in Figure \ref{fig:dataset_overview}. To construct this, we use open-source news image captioning datasets: VisualNews \cite{Liu2021-ll} and NYTimes800K \cite{Tran2020-hc}. For effectively testing caption comprehension of image-text encoders, we need to isolate the impact of caption structures from other factors that influence the synthesized image quality. NE features such as faces of specific people can pose a significant challenge to generators \cite{Rombach2022-ci}, \cite{Ramesh2022-kc}. This is likely due to the complexity of learning NE features compared to more generic visual concepts during pre-training. To assess if artifacts generated by these models are due to poor understanding of implicit context or specific entity features, we split our data into 2 distinct subsets: {\n} Non-Entity \& {\n} Entity. Similar to other internet-sourced datasets \cite{Sharma2018-tr}, we remove 95\% of low-quality image-caption pairs from a combined 1.8M samples while pre-processing as described in Appendix Section \ref{sec:dataset-preprocessing}.

\subsection{{\n} Non-Entity Subset}
This subset contains 72692 samples selected from articles published by 5 different news media organizations.  The train / val / test split of the dataset is in the ratio 90\% / 5\% / 5\% respectively. We ensure that the associated images do not contain representations of NEs in order to evaluate the influence of different caption components independently. Using a RetinaFace-based \cite{Deng2020-ht} face detector, we flag images containing identifiable faces.

\subsection{{\n} Entity Subset}
With NEs being a critical component of news image-caption pairs, this subset has been designed to evaluate the impact on T2I reasoning when NEs are present along with the other identified linguistic structures. This subset contains 7516 image-caption pairs with 48 different NEs. The current subset primarily includes PERSON entities. This is due to their frequency of mentions in news media, and consistency of physical features across images. Since there is a long-tail distribution of images per NE, we construct an eval set by selecting only 50 image-caption pairs per NE for our experiments.

\paragraph*{Multi-Modal NE Grounding}
The challenge with NE mention detection is that entities can be referred to by different names according to the situation. Example: David Beckham can be referred to as: "Beckham", "David", "David Robert Joseph Beckham", etc. To avoid this ambiguity, we need to reliably link each mention to a real-world entity. We perform Multi-modal NE grounding to link each entity mention using Wikipedia as a real-world knowledge source. Using the REL Entity Linker \cite{Van_Hulst2020-zj}, we extract entity mentions from the previously selected samples and link them to their appropriate Wikipedia pages. We used a Wikipedia dump from 2019-07 as our knowledge source. Although this helped in removing erroneous mentions detected from text captions, we also need to ensure each image contains the mentioned entity. Using their linked Wikipedia pages, we download the main image and create a repository of reference images for each entity in our dataset. Since our focus is on PERSON entities, we ensure that a face is detected in each of the downloaded reference images. A FaceNet-based \cite{Schroff2015-yg} face recognition module is used to ground each image to an entity category.

\paragraph*{Face-Aware Cropping}

The non-centered nature of foreground objects in images poses a challenge for consistent image evaluation. Many photographs are taken as long shots with the entity's face in different sections of the image. To standardize these images, we crop and resize the images taking into account the target entity position. By extracting the bounding boxes of our target entity face, we calculate its centroid as a reference coordinate for cropping. We then take a fixed window crop of the entity image such that the entity centroid is aligned closely to the center of the crop. This approach of Face-aware cropping helps maximize the image area occupied by an entity and further isolates its physical features. Through this process, we can evaluate the visual features of different entities.

\begin{figure*}
      \centering
      \includegraphics[width=\textwidth]{"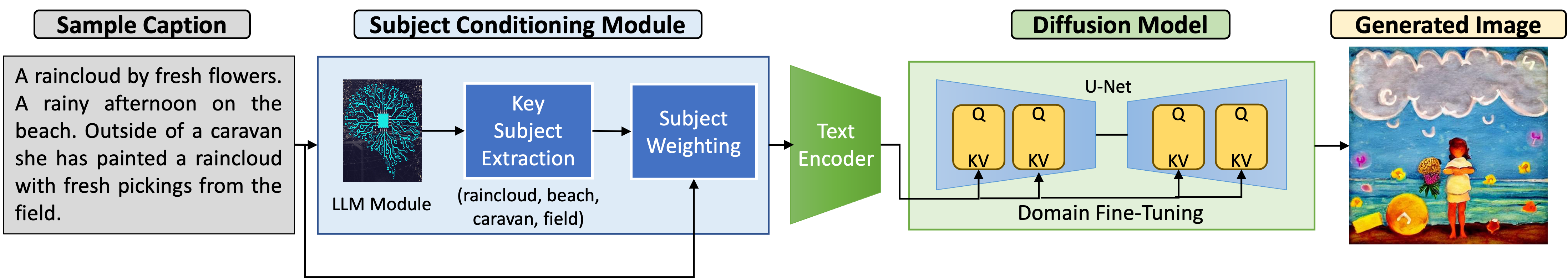"}
      \caption{Overview of our Subject-Aware FinE-tuning Approach ({\s}). Through the Subject Conditioning Module and Domain Fine-tuning, we enable better comprehension of context information in abstractive captions.}
      \label{fig:as2g_arch}
\end{figure*}

\subsection{Dataset Quality and Diversity}

We show how lexically and semantically diverse the captions in {\n} are compared to other popular datasets through quantitative metrics such as token diversity and CLIPScore in Appendix Section \ref{sec:dataset-statistics}. To further assess the overall quality of the dataset and the number of image-caption pairs that are abstractive, we conducted a human evaluation study on Amazon MTurk tabulated in Table \ref{table:human_eval_abstractive}. We consider a random sample of 300 samples extracted from the test split of {\n}. Out of the 300 selected images, 200 belong to the non-entity subset and 100 belong to the entity subset. Using this extracted sample, we perform a human evaluation of our dataset quality. The two questions we mainly aim to answer through this evaluation are: (1) Are the image-caption pairs closely related to each other from a human perspective? and (2) Are these captions Abstractive in nature? We launched our survey with 150 unique participants and each participant rated 10 samples. Per Image-caption pair, we collect 5 responses amounting to a total of 1500 responses.

\begin{table}[!ht]
\centering
    \resizebox{\columnwidth}{!}{
        \begin{tabular}{@{} c|*4c @{}}
    \toprule
    \textbf{Name} & \textbf{Size} & \textbf{Related} & \textbf{Abstractive}  & \textbf{Descriptive}\\ 
    \midrule
    {\n} Full & 300 & 291 & \textbf{260} & 31 \\
    {\n} Entity & 100 & 99 & \textbf{93} & 6 \\
    {\n} Non-Entity & 200 & 192 & \textbf{167} & 25 \\
    \hline
    \end{tabular}
    }
    \caption{Human Evaluation of {\n} Dataset Quality. Our studies show that {\n} contains highly related captions with a majority of them being abstractive in nature.}
    \label{table:human_eval_abstractive}
\end{table}

We observe that 97\% of surveyed samples are related to each other. We also see that 89.3\% of the related captions are rated as Abstractive in nature containing atleast one of the identified caption components. This supports our hypothesis and validates that our dataset pre-processing pipeline produces high-quality image-caption pairs.

\section{Understanding Abstractive Captions with External Knowledge}
\label{sec:text-cond}

At the most basic level, a subject is defined as an individual element or a concept that provides either visual, situational or complementary meaning within a caption \cite{Cohn2003-my}. We categorize the components of an abstractive caption into 4 main types: \textit{Semantic Objects, Contextual Cues, NEs, and Syntactic Variations.} Subjects such as noun phrases that have commonly identifiable visual patterns are defined as \textit{semantic objects}. \textit{Contextual cues} inform readers about the situational information surrounding a particular caption such as the setting of the image, the events leading up to that moment, and why the image's contents are significant \cite{Song2021-ln}. \textit{NEs} are domain-specific terms that refer to real-world objects and their specific features. Commonly employed NE categories include names of people, places, organizations, etc. Finally, \textit{syntactic variation} corresponds to the different forms of expression that are achieved by manipulating the relationships between different visual objects \cite{Bugliarello2021-kq}. It helps in the comprehension of complex sentences with multiple subjects and connecting clauses. 

\begin{table*}[!h]
\centering
    \begin{tabularx}{\linewidth}{@{} c|*3Y @{}}
    \toprule
    \textbf{Model} & \textbf{FID\textsubscript{CLIP}} (↓) & \textbf{IR} (↑) & \textbf{HPS V2} (↑)\\ 
    \midrule

    {\s} (DFE + GPT-3.5) & \underline{7.2804} & \textbf{0.0664} & \textbf{0.2393}\\ 

    \midrule
    \midrule
    Stable Diffusion 2.1 (CR) & 10.6595 & -0.3388 & 0.2101\\

    Stable Diffusion 2.1 (LoRA) & \textbf{6.9222} & -0.0861 & 0.2329\\

    Stable Diffusion 2.1 (Base) & 7.4780 & \underline{0.0251} & \underline{0.2385}\\
    Stable Diffusion 1.5 (Base) & 7.4742 & -0.0925 &0.2188 \\
    
    \bottomrule
    \end{tabularx}
   \caption{Results of Abstractive Text-to-Image synthesis on {\n} Non-Entity Subset. Images generated using {\s} show higher image-caption alignment while not sacrificing image fidelity compared other baselines.}
    \label{table:metrics_non_entity}
 \end{table*}

Descriptive prompts primarily utilize semantic objects with minimal presence of syntactic variations, contextual cues, and NEs while abstractive captions from domains such as news media take full advantage of all these features \cite{Zhou2022-rj, Anantha-Ramakrishnan2025-je}. For the task of T2I generation, we utilize conditioning to conform the generated image to certain criteria provided in the text input. Text conditioning in T2I models is accomplished through embeddings extracted by text encoders such as CLIP. Encoders typically employ self-attention to analyze and assign importance scores to individual elements of the input sequence while retaining the global context information across the entire sequence. This simplifies the task of word importance estimation during the generation process with every word contributing directly towards a visual concept. Let $S_{desc}$ correspond to a descriptive caption, we can represent it as a collection of subject tokens $T_i$ where $i = 1, \dots, m; m \in Z$. 

\begin{equation}
    S_{desc} = \{T_1, T_2, \dots, T_i, \dots, T_m \}
\end{equation}
 
Every subject is expected to either define or describe the properties of a visual concept present in the generated image. When generating embeddings $E_{desc}$ for a caption, each subject token is accompanied by a weight co-efficient $\alpha_i$. 
\begin{equation}
    \begin{split}
        E_{desc} & = TextEnc(S_{desc}) \\
            & = \Sigma_{i=1}^m \alpha_i*T_i \\
    \end{split}
\end{equation}

In the case of abstractive captions $S_{abstr}$, the goal of a T2I generator is to identify the salient subjects describing image contents while incorporating context information selectively. This becomes a particularly challenging task when there are multiple subjects present in the caption. Certain subjects may get forgotten or misrepresented in the generated image. This problem of "Prompt Following" \cite{Betker_undated-bq} has been documented with real-world captions. To boost comprehension of T2I generators, we aim to explicitly modify the weights $\alpha_i$ for subject tokens $T_i$. Since current encoders can capture all the relevant subjects from an image caption, we attempt to increase the weights of key subjects $T_{key}$ describing the main components of an image. Let $W_{abstr}$ correspond to the vector containing the scale multiplier $\beta > 1$ for each token. The embeddings for abstractive captions $E_{abstr}$ will be conditioned using the scale multipliers found in $W_{abstr}$. This scale multiplier helps align embeddings toward the intended meaning of a caption by acting as a prompt weight. 

\begin{equation}   
    W_{abstr} = 
         \begin{cases}
           \text{$\beta$,}  &\quad\text{if $T_i$} \in T_{key}\\
           \text{1,} &\quad\text{otherwise.} \\
         \end{cases}
\end{equation}

\begin{equation}
    E_{abstr} = TextEnc(S_{abstr}) \odot W_{abstr}
\end{equation}

\subsection{Subject-Aware FinE-tuning ({\s})}
As illustrated in Figure \ref{fig:as2g_arch}, {\s} utilizes a Stable Diffusion-based backbone for T2I generation. By taking advantage of prompt weighting and fine-tuning strategies, our approach aims to enhance both the image fidelity and prompt following capabilities over baselines.

\begin{figure*}[t]
\centering
\begin{tabular}{cccc}
\hline
 \textbf{Original} & \textbf{SD 2.1 (Base)} & \textbf{SD 2.1 (CR)} & \textbf{{\s} (DFE + GPT-3.5)} \\ \toprule
\makebox[0.21\textwidth]{\includegraphics[width=0.15\textwidth]{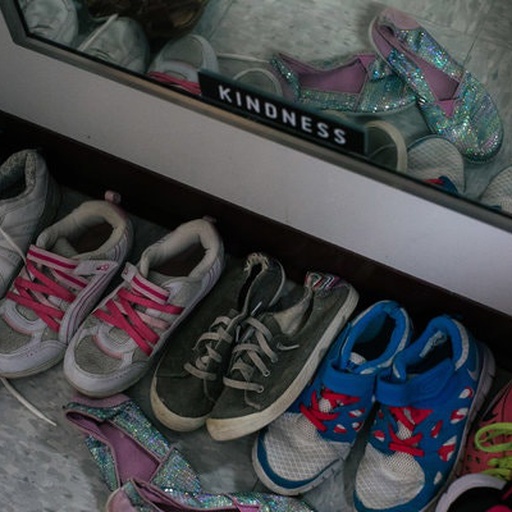}} & \makebox[0.21\textwidth]{\includegraphics[width=0.15\textwidth]{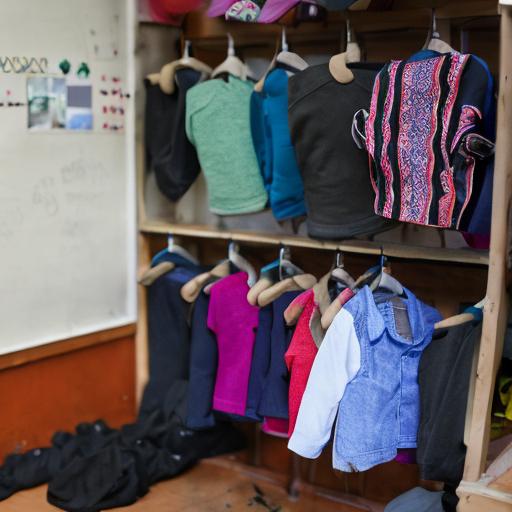}} & \makebox[0.21\textwidth]{\includegraphics[width=0.15\textwidth]{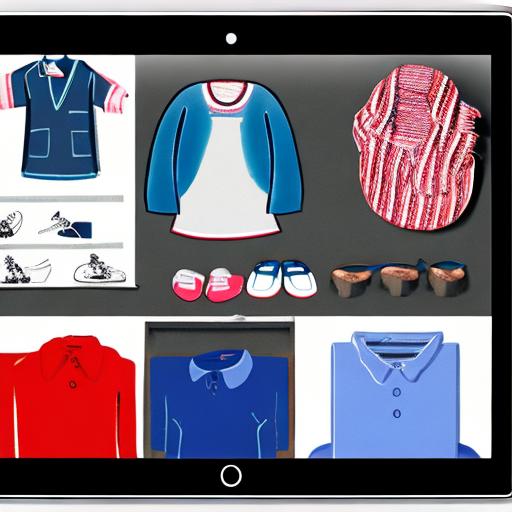}} &
\makebox[0.21\textwidth]{\includegraphics[width=0.15\textwidth]{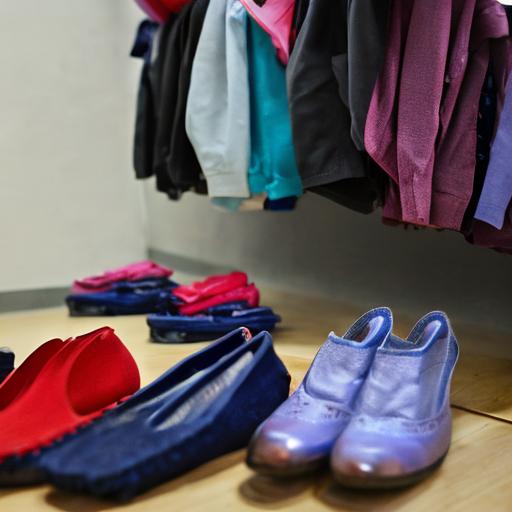}}\\

& \textcolor{red}{(Missing Shoes)} & \textcolor{red}{(Misleading Style)} & \textcolor{MySoftGreen}{(Shows shoes \& clothes)} \\

\multicolumn{4}{p{0.95\textwidth}}{\centering \hypertarget{fig:gen_results1}{\textbf{Ex1:} The school offers \textcolor{orange}{clothing}, including \textcolor{orange}{shoes}, to its \textcolor{orange}{students}. }} \\ 
& & & \\

\noalign{\vskip -7pt}
\midrule

\makebox[0.21\textwidth]{\includegraphics[width=0.15\textwidth]{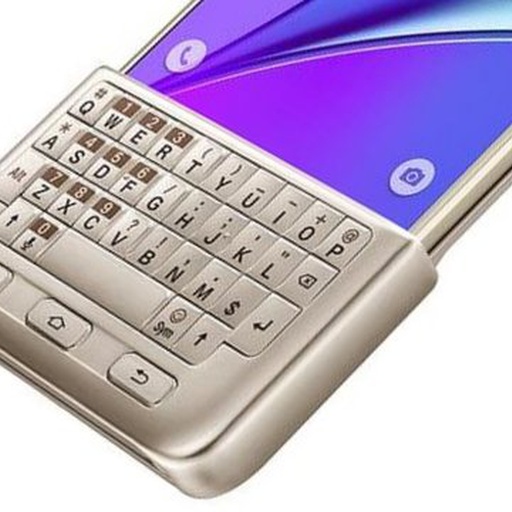}} & \makebox[0.21\textwidth]{\includegraphics[width=0.15\textwidth]{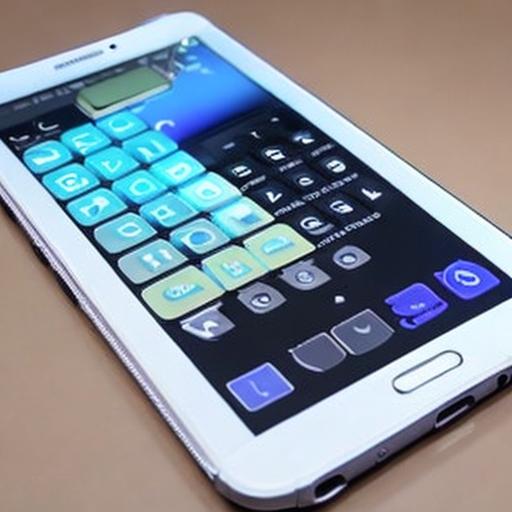}} & \makebox[0.21\textwidth]{\includegraphics[width=0.15\textwidth]{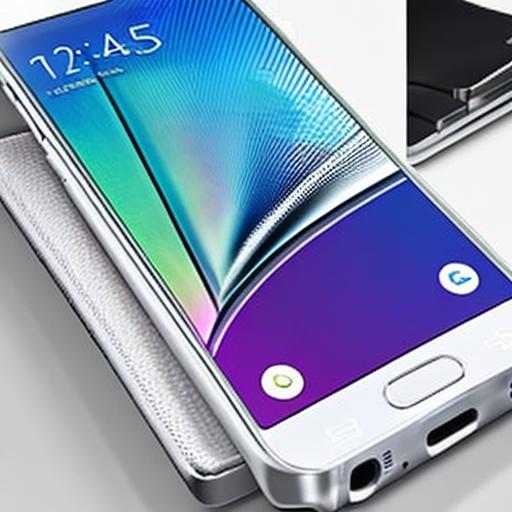}} & \makebox[0.21\textwidth]{\includegraphics[width=0.15\textwidth]{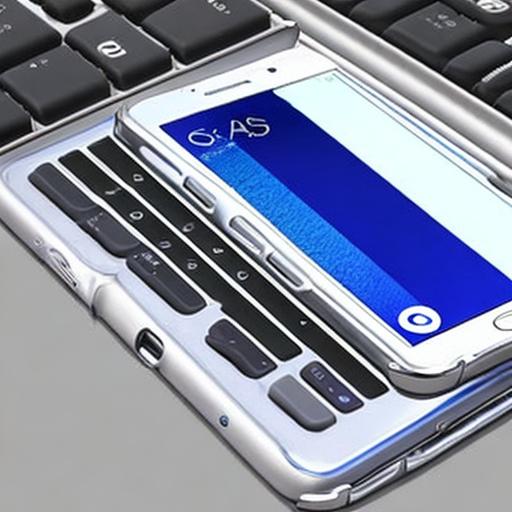}}\\ 

& \textcolor{red}{(Missing Case)} & \textcolor{red}{(Missing Keyboard)} & \textcolor{MySoftGreen}{(Shows case with keyboard)} \\

\multicolumn{4}{p{0.95\textwidth}}{\centering \hypertarget{fig:gen_results3}{\textbf{Ex2:} The \textcolor{orange}{Galaxy Note 5} can be used with a \textcolor{orange}{case} that doubles as a physical \textcolor{orange}{Qwerty keyboard} to aid \textcolor{orange}{typing}. }} \\ 
& & & \\

\noalign{\vskip -7pt}
\midrule

\makebox[0.21\textwidth]{\includegraphics[width=0.15\textwidth]{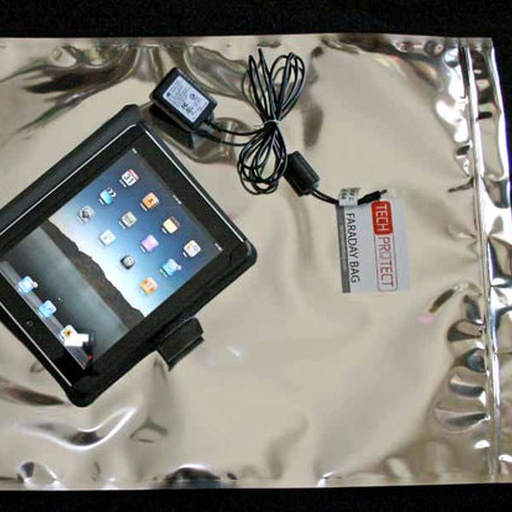}} & \makebox[0.21\textwidth]{\includegraphics[width=0.15\textwidth]{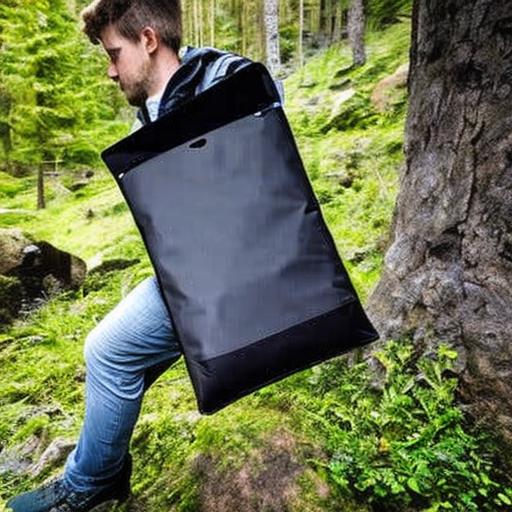}} & \makebox[0.21\textwidth]{\includegraphics[width=0.15\textwidth]{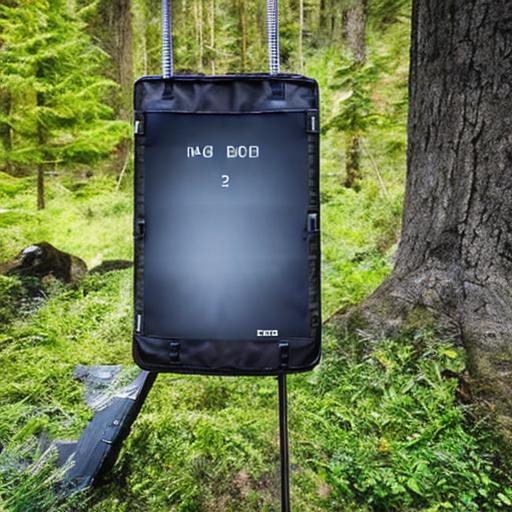}} & \makebox[0.21\textwidth]{\includegraphics[width=0.15\textwidth]{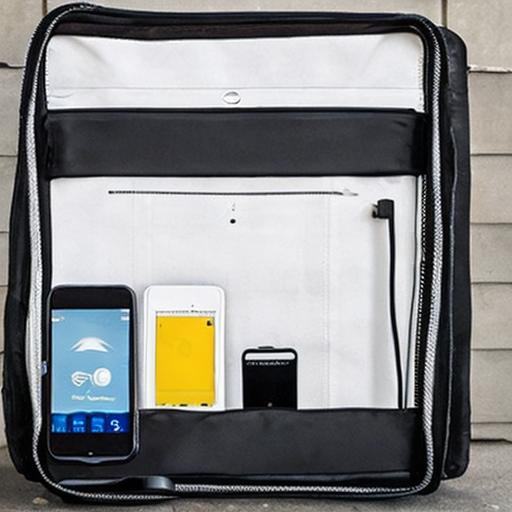}}\\

& \textcolor{red}{(Incorrect Human Subject)} & \textcolor{red}{(Incorrect Context)} & \textcolor{MySoftGreen}{(Shows correct context)} \\

\multicolumn{4}{p{0.95\textwidth}}{\centering \hypertarget{fig:gen_results5}{\textbf{Ex3:} 
A \textcolor{orange}{Faraday bag}, which blocks \textcolor{orange}{remote signals} to \textcolor{orange}{devices} such as \textcolor{orange}{cellphones} and \textcolor{orange}{tablets}. 
}} \\

\end{tabular}
    \caption{Qualitative comparison of different T2I models on {\n} Non-Entity. Words highlighted in \textcolor{orange}{Orange} are used for subject weighting in {\s}. Incorrect Visual Artifacts in images are described in \textcolor{red}{Red} and the correct contextual expectations for generated images are marked in \textcolor{MySoftGreen}{Green}.}
    \label{figure:gen_results}
\end{figure*}

\paragraph*{LLMs for Subject Conditioning} 
The key challenge in implementing subject conditioning is identifying which subjects/phrases to weigh positively. To replace human guidance in the process of prompt weighting, we evaluate the use of LLMs in identifying salient subjects from sentences. Leveraging the commonsense reasoning abilities of LLMs, we utilize instruction-based prompting to extract salient subjects from each sentence. Subject identification is done in a zero-shot manner using only the prompt and the pre-trained world knowledge of LLMs. This process allows us to explicitly condition the input embeddings in an observable and explainable manner. Compared to other LLM-based grounding strategies \cite{Lian2024-df, Feng2023-mk}, subject conditioning requires only single-stage prompting and also utilizes fewer tokens per generation. Additionally, we also compare different LLM architectures including both Commercial (GPT-4, GPT-3.5) and Open-source (Mixtral 8x7B, Orca Mini-13B)  \cite{Jiang2024-ht, Mukherjee2023-ye} models on this task as described in Appendix Section \ref{sec:salient-subject-selection}. This comparative study helps us ascertain the types of models capable of producing high-quality subject weights.

\paragraph*{Handling Domain-Shift}
Real-world images and captions have specific characteristics that differ from the general outputs generated by T2I models. Specifically, real-life photographs with specific foreground and background objects are present in abundance compared to artistic or animated-style images. To tackle this domain shift, we develop our Domain fine-tuning (DFE) strategy on {\n}. Traditional Mean Squared Error-based losses used for Stable Diffusion fine-tuning tend to generate unrealistic images \cite{Zhang2018-zo, Lin2023-sz}, making them unsuitable for our task. We adopt the Reward Feedback Learning (ReFL) \cite{Xu2023-uo} strategy for directly optimizing Stable Diffusion on a reward model trained to score image-caption alignment. The selected reward model ImageReward (IR) \cite{Xu2023-uo} has been trained on 137K human-annotated samples to predict alignment between image-caption pairs. Our proposed improvements in DFE over vanilla ReFL focus on improving both alignments with the ground truth image and caption, instead of only caption alignment. In DFE, we initialize the latent vector of Stable Diffusion based on the ground truth images for each caption instead of random initialization as implemented in ReFL. This helps control the diffusion process in generating target distribution-aligned images. To increase training stability, we fine-tune only the Attention Layers using Low Rank Adaptation (LoRA) \cite{Hu2022-kw}. Additionally, the noise scheduler of the original ReFL pipeline was limited to having 40 timesteps. The authors identified that latents sampled between timesteps 30-39 produced distinguishable IR scores. We extend this by our noise scheduler to 100 timesteps and sampling from steps 40-99 for loss calculation. 
\section{Experiments and Results}
\label{sec:experiemnts}

\subsection{Evaluation Metrics}

To holistically evaluate the samples generated by T2I models on {\n} Non-Entity, we report 3 different types of metrics: CLIP-based Frechet Inception Distance ($FID_{CLIP}$) \cite{Kynkaanniemi2022-de}, IR \cite{Xu2023-uo}, and Human Preference Score (HPS) V2 \cite{Wu2023-nm}. On {\n} Entity, we include Identity Preservation and Face Detection accuracy as additional metrics to quantify their entity image generation performance similar to \cite{Yuan2023-oj}. We provide additional details on metric selection and model hyperparameters in Appendix Section \ref{sec:implementation-details}.

\subsection{Baseline Models} 
Our goal is to study \textit{encoder-level} behavior in a controlled setting, thus we compare single-encoder models primarily as our baselines. Newer architectures that rely on multiple text encoders, LLM-based encoders, and adapter/fusion layers mix several embedding streams, making encoder-level analysis difficult \cite{Toker2024-al, Gao2024-fj}. Thus we select: Stable Diffusion V1.5 and V2.1 \cite{Rombach2022-ci} as representative models for T2I generation. These models utilize CLIP-based architectures for image-text encoders, showing strong performance in traditional benchmarks such as COCO Captions \cite{Chen2015-qj}. Additionally, we also compare the performance of LLM-powered Caption Rewriting (CR) for translating abstractive captions into descriptive text. Specifically, we use an LLM to rewrite a caption into an instruction prompt of the format "Generate an image ...".

\subsection{Quantitative Results}
On analysis of the scores presented in Table \ref{table:metrics_non_entity}, {\s} outperforms both baseline Diffusion models on the {\n} Non-entity test set. Similarly, LoRA fine-tuning only improves on image fidelity, while showing lower image-caption alignment. The explicit guidance through both fine-tuning and subject conditioning contributes towards capturing the intended meaning of captions and also producing more aligned images without changing the encoder architecture. Caption rewriting on the other hand performs significantly worse on our benchmark metrics, signaling that CLIP-based encoders face challenges in comprehending situated context even when translated descriptively.

For {\n} Entity, we report the average metric scores across all entity classes in Appendix Table \ref{table:metrics_entity}. Our results show that captions with NEs are highly sensitive to visual degradation upon embedding re-weighting \cite{Xiao2025-yg}, as evidenced by the overall lower Detect and Identity scores. Although multiple entities may be better represented through re-weighting, the overall quality of the generated image is impacted adversely, demonstrating the challenging nature of this problem. This justifies our development of two subsets within {\n} to support this analysis.

\subsection{Qualitative Results}
From the examples presented in Figure \ref{figure:gen_results}, we observe that {\s} achieves more faithful prompt interpretation than the baselines by effectively handling contextual cues and syntactic variations within captions. In the first example, the subjects “clothing”, “shoes”, and “students” are semantically related but require correct syntactic parsing to preserve their relationships. {\s} successfully captures these dependencies, producing images with visible shoes and clothing. In contrast, SD 2.1 (Base) fails to represent all key subjects, particularly the shoes. SD 2.1 (CR) generates a more grounded image but misinterprets the contextual framing, rendering the scene as a sketch rather than a realistic photo. This reflects how caption rewriting can distort the original style cues embedded in the caption structure. In the second example, a contextual focus is placed on the “keyboard”, requiring the model to understand both the object and its role within the scene. SD 2.1 (Base) again overlooks this nuance, producing an image where the keyboard is missing. {\s}, however, correctly maintains the contextual intent and learns the syntactic variation of the caption, resulting in a composition that aligns closely with the described scenario. Human evaluation results in Table \ref{table:qualitative_results} further confirm that images generated by {\s} are consistently preferred for their stronger contextual grounding and structural coherence. Details regarding study setup and design are provided in Appendix \ref{sec:qual-eval-setup}.

\begin{table}[!ht]
    \centering
    \resizebox{\columnwidth}{!}{
        \begin{tabular}{@{} c|*3c @{}}
        \toprule
        \textbf{Model} & \textbf{Total} & \textbf{Preferred Samples} & \textbf{Preference (\%)} \\ 
        \midrule
        {\s} (DFE + GPT-3.5) & 185 & 102 & \textbf{55.13}\\
        Stable Diffusion 2.1 & 185 & 83 & 44.86 \\ 
    
        \bottomrule
        \end{tabular}
        }
       \caption{Human Evaluation Results on {\n}. We show that there exists higher preference among users for images generated by {\s} compared to T2I baselines.}
        \label{table:qualitative_results}
\end{table} 

\begin{table}[!ht]
    \centering
    \resizebox{\columnwidth}{!}
    {
        \begin{tabular}{@{} c|*3c @{}}
        \toprule
        \textbf{Model} & \textbf{FID\textsubscript{CLIP}} (↓) & \textbf{IR} (↑) & \textbf{HPS V2} (↑)\\ 
        \midrule
    
        {\s} (GPT-3.5) & \underline{7.2614} & 0.0624 & \underline{0.2392}\\
        {\s} (GPT-4) & \textbf{7.2482} & \underline{0.0673} & 0.2391\\
        {\s} (Mixtral 8x7B) & 7.2649 & \textbf{0.0723} & \textbf{0.2394} \\
        {\s} (Orca Mini-13B) & 7.3571 & 0.0298 & 0.2381 \\

        \bottomrule
        \end{tabular}
    }
    \caption{Ablation study evaluating the impact of different LLM Agents for subject weight extraction. Open Source LLM backbones perform equally well when compared to proprietary models for subject weighting.}
    \label{table:ablation_study_llm}
\end{table}

On {\n}'s Entity Subset, the generated images using {\s} cohesively include most objects presented in the caption, while the images generated by SD 2.1 (Base) focus primarily on the NEs present. As shown in the Appendix Figure \ref{figure:suppl-ent-gen-results}, although SD 2.1 (Base) does have challenges such as the repeatedly generating the same entity, the overall quality of NE features is higher. This presents itself as better alignment across metrics. 

\subsection{Ablation Study}

\paragraph*{Subject Weight Quality Across LLM Models}
To assess the variation in commonsense reasoning and world knowledge of different LLM architectures, we collect subject weights from 4 different LLMs: GPT-3.5, GPT-4, Orca Mini-13B, and Mixtral 8x7B Mixture of Experts (MoE). For our ablation study, we replace only the provided subject weights from each model during inference using SD 2.1 (Base) as our T2I backbone as shown in Table \ref{table:ablation_study_llm}. We can observe a clear correlation between LLM performance on other commonsense reasoning tasks and key subject delineation with Mixtral and GPT-4 outperforming other models. Subject weights generated by models with lower number of parameters like Orca Mini-13B still show improvements on 2 of the 3 metrics compared to our baselines. This demonstrates the potential of open-source LLMs in boosting caption understanding for cross-modal generative tasks. 

\begin{table}[H]
    \centering
    \resizebox{\columnwidth}{!}
    {
        \begin{tabular}{@{} c|*3c @{}}
        \toprule
        \textbf{Model} & \textbf{FID\textsubscript{CLIP}} (↓) & \textbf{IR} (↑) & \textbf{HPS V2} (↑)\\ 
        \midrule
        {\s} (DFE + GPT 3.5) (x\textsuperscript{2}) & \underline{7.2804} & \textbf{0.0664} & \underline{0.2393}\\
        \midrule
        w/o GPT-3.5 & 7.4851 & 0.0249 & 0.2385\\
        w/o DFE & \textbf{7.2614} & \underline{0.0624} & 0.2392\\
        \midrule
        \midrule
        {\s} (DFE + GPT 3.5) (x\textsuperscript{1}) & 7.3729 & 0.0564 & \textbf{0.2395}\\
        {\s} (DFE + GPT 3.5) (x\textsuperscript{3}) & 7.3049 & 0.0040 & 0.2361\\
        {\s} (DFE + GPT 3.5) (x\textsuperscript{4}) & 7.8825 & -0.1835 & 0.2255\\

        \bottomrule
        \end{tabular}
    }
    \caption{Ablation study evaluating the effectiveness of different scale multiplier values (x\textsuperscript{n}) and model components. }
    \label{table:ablation_study_subweight}
\end{table}

\paragraph*{Impact of Subject Conditioning} 
\label{sec:ablation-subject-conditioning}

We investigate the impact of each of {\s}'s components on generation quality as shown in Table \ref{table:ablation_study_subweight}. We observe that Subject Conditioning provides a significant contribution towards the observed metric performance. The addition of DFE also boosts image-caption understanding without majorly impacting image fidelity, as reflected in all 3 metrics presented. The positive correlation between IR and HPS V2 even with the addition of DFE, confirms that the fine-tuning process hasn't overfit on the reward model.

\paragraph*{Impact of Subject Scale Multiplier}
\label{sec:ablation-subject-scale}

We evaluate various candidate score multipliers as shown in Table \ref{table:ablation_study_subweight}. Here, x\textsuperscript{1} refers to a scale factor of $1.1$, x\textsuperscript{2} refers to a scale factor of $(1.1)^2$, and so forth. We selected a multiplier of x\textsuperscript{2} as it scores the highest in 2 out of 3 metrics tested. Increasing it beyond x\textsuperscript{2} does not provide any meaningful improvements.

\paragraph*{Generalizability of Subject Weighting across T2I Benchmarks}
\label{sec:anchor-existing-bench}

To understand if our proposed subject weighting methodology is generalizable to other task domains apart from news image synthesis, we evaluate our approach on existing benchmark T2I datasets. We selected the Conceptual Captions (CC3M) dataset \cite{Sharma2018-tr} as a relevant benchmark. Since CC3M is sourced from web-scraped articles and blog posts, it offers a comparable level of caption complexity to {\n}. Our experiments cover both baseline and SAFE models on the public validation set of CC3M as described in Table \ref{table:cc3m-results}. We show that our proposed weighting approach improves context relatedness and human preference alignment, even without any domain fine-tuning. 

\begin{table}[!h]
    \centering
    \resizebox{\columnwidth}{!}
    {
        \begin{tabular}{@{} c|*3c @{}}
        \toprule
        \textbf{Model} & \textbf{FID\textsubscript{CLIP}} (↓) & \textbf{ImageReward} (↑) & \textbf{HPS V2} (↑)\\ 
        \midrule
        SAFE (GPT 3.5) & 8.4124	& \textbf{0.2700} & \textbf{0.2521}\\
        Stable Diffusion 2.1  & \textbf{8.3883} & 0.2414 & 0.2516\\
        \bottomrule
        \end{tabular}
    }
    \caption{Ablation Results of T2I synthesis on CC3M dataset. We show the extendability of {\s} to other T2I benchmarks that contain Non-Entity captions.}
    \label{table:cc3m-results}
\end{table}
\section{Conclusions}

We present {\n}, a novel dataset for evaluating image-text encoders such as CLIP on abstractive captions, identifying key challenges in multi-subject understanding and context-based reasoning. To mitigate this, our subject conditioning strategy: {\s} helps improve subject grounding and interpreting syntax variations. With {\s} being able to re-rank the importance of specific subjects at an embedding-level, we improve contextual alignment without increasing parameter size or retraining the encoder. Through fine-tuning, we integrate both image-level and human-preference alignment objectives, building on top of traditional techniques such as LoRA and ReFL. Compared to other LLM + Diffusion methods \cite{Liao2024-un, Lian2024-df}, {\s} requires only one LLM query with significantly fewer tokens returned, lowering inference cost and processing time.

\section*{Limitations}
Since we build on top of open-source Large Foundational Models such as Stable Diffusion, GPT-3.5, GPT-4, Mixtral-8x7B, and Orca, our approach inherits all their biases. We do not analyze T2I models that use multi-encoder architectures \cite{Esser2024-ha, Chen2023-ni} given the challenges in disentangling the influence of individual encoders towards multi-subject and context comprehension. Extending SAFE to multilingual abstractive captions also faces several challenges in validating the abstractiveness of publicly available datasets and the inherent weakness of multilingual CLIP variants in performing contextual grounding on low-resource languages \cite{Ananthram2024-eb}. The lack of task-specific fine-tuning to improve entity likeness generation is another limitation that our approach faces. Future research directions include development of entity concept datasets and analyzing unified multi-modal and multi-lingual encoders that learn on both image and text tokens simultaneously. 

\section*{Ethical Considerations}
We only source data from publicly available news media repositories that are licensed for use in research. We will also release our dataset under the same license restrictions for public access (CC BY-NC-SA 4.0). To remove NSFW content, we apply a combination of profanity word-lists \cite{NguyenUnknown-bg} and machine learning-based caption filtering methods \cite{ZhouUnknown-rt}. Our goal for launching a news media-focused abstractive captions dataset is solely to evaluate the quality and alignment of image-text embeddings used for conditional guidance. However, to mitigate any potential risk of these models being used for misinformation generation, we recommend strict guidelines on the responsible use of this technology. This includes using the model only for illustrative purposes and not for creating images that represent real-world events without human oversight. We used LLM-based AI-based proofreading tools solely for minor language and grammar corrections after completing the scientific content. These tools were not used for ideation, writing, analysis, or data generation. All intellectual contributions are those of the authors.

\section*{Acknowledgments}
Special thanks to Aadarsh Anantha Ramakrishnan for his contributions through insightful research discussions and proofreading of the draft. This research has been partially supported by NSF Awards \#1820609 and \#2114824.

\bibliography{paperpile}

\clearpage

\appendix

\section*{Appendix}
\label{sec:appendix}

\section{Dataset Pre-processing}
\label{sec:dataset-preprocessing}

\subsection{Image-based Filtering}
We standardize the resolution of all images to 512x512. By using Entropy-based cropping, we retain focus on points of interest within a frame. This helps keep the foreground object at the center of the image, limiting information loss to only the background elements. To remove noisy and blurry images, we use CLIP-IQA \cite{Wang2023-mr} as a reference-free metric. To filter images based on noisiness and sharpness, we use a minimum threshold of 0.3.

\subsection{Caption Filtering and Entity Tagging}
In the first stage of filtering, we remove very short captions. We select captions with a minimum length of 6 words and above for our dataset. This is done to ensure that selected captions are informative enough for T2I synthesis. We use different approaches for tagging NEs based on the dataset the captions were extracted from. For captions extracted from the NYTimes800K news corpus, we use the provided NER annotations for filtering. The VisualNews corpus does not provide ground-truth annotations, so we identify mentions of NE using the Spacy library. We remove samples containing 'PERSON', 'GPE', 'LOC', 'WORK\_OF\_ART', 'ORG' entity types due to their high presence in captions.

\begin{table*}[h!]
\centering

    \begin{tabularx}{\linewidth}{@{} l| Y | Y | Y | Y @{}}
    \toprule
    
    \multirow{2}{*}{\bfseries Dataset} & 
    \multirow{2}{*}{\bfseries Unique Tokens} & 
    \multirow{2}{*}{\bfseries CLIPScore (↑)} &
    \multicolumn{2}{c}{\bfseries Caption Length}\\ \cmidrule{4-5}
    & & & \textbf{Mean} & \textbf{StdDev}\\ \midrule
    COCO Captions Train & 22767 & 0.5152 & 10.42 & 0.88\\
    COCO Captions Val & 16647 & 0.5237 & 10.42 & 0.87 \\
    \midrule
    CC3M Train & 45896 & 0.3984 & 10.31 & 3.30 \\
    CC3M Val & 9289 & 0.4880 & 10.40 & 3.35 \\
    \midrule
    \midrule
    {\n} Non-Entity Train & 51026 & 0.4797 & 14.84 & \underline{5.51} \\
    {\n} Non-Entity Val & 10619 & 0.4811 & \underline{14.93} & 5.38 \\
    {\n} Non-Entity Test & 10485 & 0.4807 & 14.67 & 5.36 \\
    \midrule
    {\n} Entity & 10955 & 0.4960 & \textbf{22.13} & \textbf{7.95} \\
    \bottomrule
    \end{tabularx}

   \caption{Caption Statistics of {\n}. Overall, we observe the captions from {\n} are more lexically and semantically diverse compared to traditional T2I evaluation benchmarks.}
    \label{table:dataset-stats}
 \end{table*}

 \section{Dataset Insights}
\label{sec:dataset-statistics}

\subsection{Caption Statistics}

In this section, we provide additional statistics on the {\n} dataset and analyze the distribution of image-caption pairs. In Table \ref{table:dataset-stats}, we provide caption statistics of {\n} compared to 2 popular image-caption pair datasets: COCO Captions \cite{Chen2015-qj} and Conceptual Captions 3M (CC3M) \cite{Sharma2018-tr}. By tokenizing and lemmatizing each caption without case sensitivity, we compute the number of unique tokens present in a dataset. We utilize the NLTK library for both tokenization and lemmatization. We can observe that across different data splits of {\n}, the mean caption length is significantly higher with a greater variation in caption length compared to other datasets. In addition to the increased caption length, it also contains a significant amount of unique tokens considering the number of samples present. The CLIPScores of samples in {\n} are also lower compared COCO Captions, indicating that default CLIP embeddings may struggle in capturing contextual alignment in abstractive captions. This highlights the diversity of captions in {\n}, showing greater expression in describing visual concepts.

\begin{figure*}
      \centering
      \includegraphics[width=0.8\linewidth]
      {"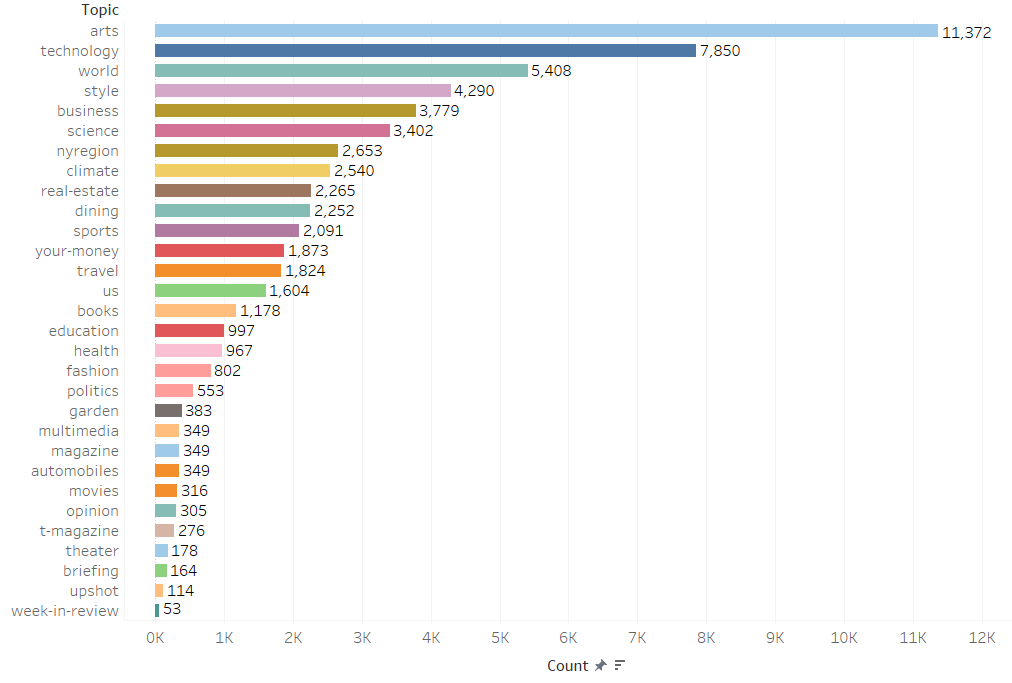"}
      \caption{Distribution of Article Topics for samples in {\n}.}
      \label{fig:article_topics}
\end{figure*}

\begin{algorithm}
\caption{Subject Reweighting Process}
\label{alg:subject_reweighting}
\KwIn{Abstractive Caption $S_{abstr}$, Large Language Model LLM, Tokenizer $\mathcal{T}$, Text Encoder $TextEnc$, T2I Model $\mathcal{G}$, Scale Multiplier $\beta$}
\KwOut{Generated Image $I$}

\vspace{2mm}
\textbf{Step 1: Tokenization}\\
$\{T_1, T_2, \dots, T_m\} \gets \mathcal{T}(S_{abstr})$ \tcp*{Tokenize caption into subject tokens $T_i$}

\vspace{2mm}
\textbf{Step 2: Identify Subject Spans}\\
$T_{key} \gets \text{LLM}(\{T_1, \dots, T_m\})$ \tcp*{Pass tokens to the LLM to pick key subjects}

\vspace{2mm}
\textbf{Step 3: Assign Weights}\\
Initialize weight vector $W_{abstr}$ of size $m$ with $1$\;
\For{$i = 1$ \KwTo $m$}{
    \If{$T_i \in T_{key}$}{
        $W_{abstr}[i] \gets \beta$ \tcp*{Assign upweight ($\alpha_i = \beta$) to key tokens}
    }
}

\vspace{2mm}
\textbf{Step 4: Generate Original Embeddings}\\
$E_{orig} \gets TextEnc(S_{abstr})$ \tcp*{Generate the original/unmodified embeddings}

\vspace{2mm}
\textbf{Step 5: Apply Weights}\\
$E_{abstr} \gets E_{orig} \odot W_{abstr}$ \tcp*{Multiply each token embedding by its weight}

\vspace{2mm}
\textbf{Step 6: Image Generation}\\
$I \gets \mathcal{G}(E_{abstr})$ \tcp*{Use reweighted embeddings as guidance for T2I}

\Return{$I$}\;
\end{algorithm}

\subsection{Article Topic Distribution}

For analyzing the categories of articles from which image-caption pairs have been selected for our dataset, we provide a unified category list in Figure \ref{fig:article_topics}. With articles sourced from different news agencies, each source has its own article category taxonomy. To create a unified taxonomy, we fix the categories provided by articles from NYTimes as our template. To cluster similar article categories under one label, we utilize the lightweight sentence transformer all-MiniLM-L6-v2 \cite{UnknownUnknown-es}. With a minimum similarity threshold of 0.5, we cluster every sample's default topic description into NYTimes's taxonomy labels. Here, we visualize our dataset's top 30 article classes, showing the diverse spread of image-caption pairs present. 

\begin{tcolorbox}[title={GPT-3.5/4 \& Mixtral 8x7B Prompt}, boxsep=0.5pt]
\textbf{User:} Use only the information provided in the prompt for answering the question. List the main topic word and additional topic words from the given image caption in the format: \{"main\_topic\_word": \textless{}insert-topic-word-string\textgreater{}, "additional\_topic\_words": [ \textless{}insert-topic-word1\textgreater{}, ... ]\}. Caption Text: \textless{}insert-caption-text\textgreater{}
\end{tcolorbox}

\begin{tcolorbox}[title={Orca Mini 13B Prompt}, after skip=0pt, boxsep=0.5pt]
\textbf{User:} "User: List only the main objects from the sentence: \textless{}insert-caption-text\textgreater{}"
\end{tcolorbox}

\begin{tcolorbox}[title={Caption-Rewriting Prompt}, boxsep=0pt] 
\textbf{User:} Write a simple prompt for an image generation model to generate an image for the given text: \textless{}insert-caption\textgreater{}
\end{tcolorbox}

\section{Salient Subject Selection}
\label{sec:salient-subject-selection}

The detailed overview on how we perform Subject weighting using the Compel library \cite{DamianUnknown-aj} is described in Algorithm \ref{alg:subject_reweighting}. To reduce memory requirements and to speed up inference, we initialize in mixed precision mode and set $dtype = float16$. The system prompt is set as ``\textit{You are an AI assistant that follows instructions extremely well. Help as much as you can.}" In cases where the LLM returned no salient subject phrases, we use only the text caption without any subject weights. Given the size of our dataset (70K+ image-caption pairs), GPT 3.5 Turbo offered the best balance of consistency, latency, and token cost, making it our preferred backbone to run the majority of our ablation studies.

\section{Implementation Details} 
\label{sec:implementation-details}

\subsection{Metric Selection and Justification}

Frechet Inception Distance \cite{Heusel2017-ob} serves as an indicator to quantify the overall realism and diversity of generated samples compared to the ground truth images. With the distribution of datasets like {\n} diverging significantly from the Inception-V3 used in traditional FID calculations \cite{Kynkaanniemi2022-de}, we adopt the more representative $FID_{CLIP}$ metric for our testing. To measure the relatedness of our generated images and ground truth captions, we utilize IR. Compared to image-caption similarity metrics like CLIPScore \cite{Hessel2021-we}, IR is trained on real-world image-caption pairs annotated and ranked according to human preference. Similarly, Human Preference Score V2 also serves as an indicator of human preference alignment. For Face Detection, we utilize a RetinaFace-based detector and measure the average number of times a face is detected across generated images. The ArcFace \cite{Deng2019-li} face recognition model is used for calculating Identity Preservation scores.  

\subsection{Model Training and Inference}
We set \textit{guidance\_scale = 7.5} and \textit{num\_inference\_steps = 100}. All our reported metrics are averaged across 2 random seeds 42 and 3. All generated images are in 512x512 resolution. Our {\s} Model has been fine-tuned for 300 epochs. We utilize the same inference hyper-parameters as the baseline Stable Diffusion models. A learning rate of $5*10^{-5}$ was set for {\s}. All experiments were performed on a Nvidia A100 GPU taking up to 200 GPU hours to run. To perform prompt weighting, after experimenting with different weight scales, we apply a uniform increase of x2 or $(1.1)^2$ for all LLM extracted keywords in the original caption. Generated samples of Baseline and {\s} models using the same seed are presented in Figure \ref{figure:gen_results}. We select GPT-3.5 as our default LLM model for collecting subject weights for all our fine-tuned models. On the {\n} Entity test-set, we assess the impact subject conditioning has in understanding abstractive captions containing NEs.

\begin{table*}[!h]
    \centering
    \begin{tabularx}{\linewidth}{@{} c|*6Y @{}}
    \toprule
    \textbf{Model} & \textbf{Identity} (↑) & \textbf{Detect} (↑) & \textbf{FID\textsubscript{CLIP}} (↓) & \textbf{IR} (↑) & \textbf{HPS V2} (↑)\\ 
    \midrule
    {\s} (DFE + GPT-3.5) & 0.3323 & 0.9498 & 30.7756 & \textbf{0.6072} & 0.2564\\ 
   
    Stable Diffusion 2.1 (Base) & \textbf{0.3391} & \textbf{0.9533} & \textbf{30.0651} & 0.6060 & \textbf{0.2565}\\
    \bottomrule
    \end{tabularx}
   \caption{Results of Abstractive Text-to-Image synthesis on {\n} Entity Test-set averaged across all classes. As expected, the introduction of {\n} results in a drop in generation quality of NE features while retaining image-caption alignment. This supports our hypothesis that Non-Entity and Entity features are represented differently by encoders such as CLIP.}
    \label{table:metrics_entity}
 \end{table*}

\begin{figure*}[!t]
\centering
\begin{tabular}{ccc}
    \textbf{Original} & \textbf{SD 2.1 (Base)} & \textbf{SAFE (DFE + GPT-3.5)} \\ \toprule

    \makebox[0.3\textwidth]{\includegraphics[width=0.15\textwidth]{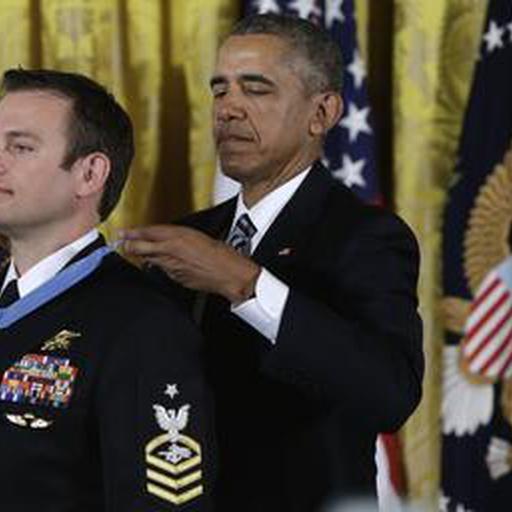}} & \makebox[0.3\textwidth]{\includegraphics[width=0.15\textwidth]{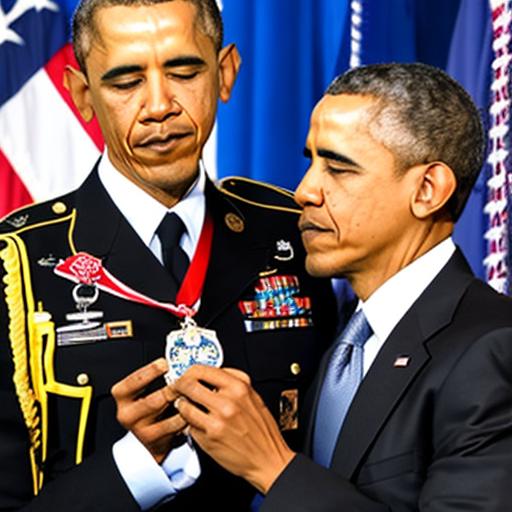}} & \makebox[0.3\textwidth]{\includegraphics[width=0.15\textwidth]{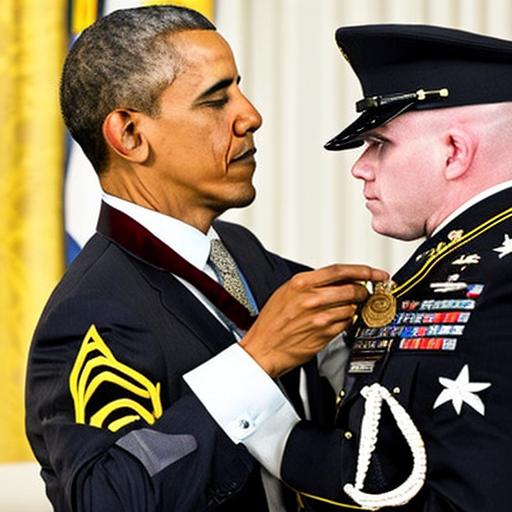}} \\

    \multicolumn{3}{p{0.95\textwidth}}{\centering \hypertarget{fig:suppl-ent-gen4}{\textbf{Ex4:} \textcolor{orange}{Obama} awards \textcolor{orange}{Medal of Honor} to member of \textcolor{orange}{SEAL Team 6}. 
    }} \\ 
    & & \\
    
    \noalign{\vskip -7pt}
    \midrule

    \includegraphics[width=0.15\textwidth]{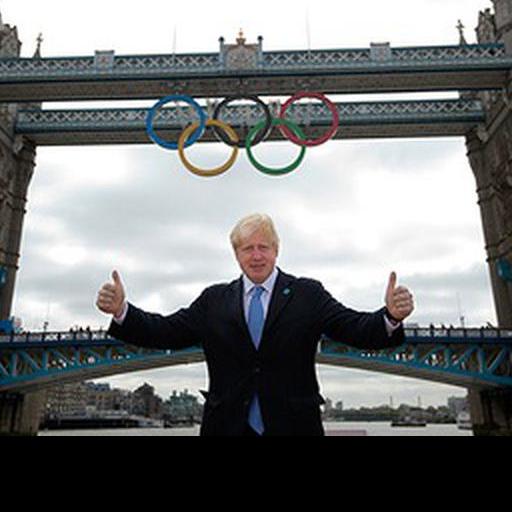} & \includegraphics[width=0.15\textwidth]{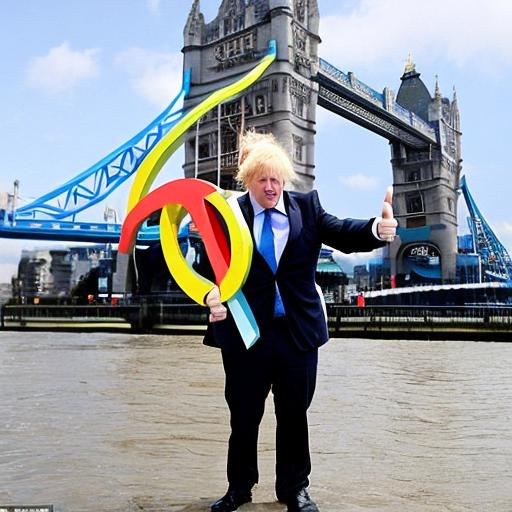} & \includegraphics[width=0.15\textwidth]{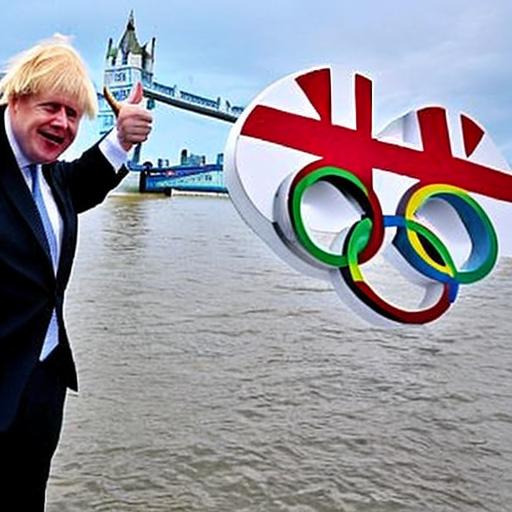} \\

    \multicolumn{3}{p{0.95\textwidth}}{\centering \hypertarget{fig:suppl-ent-gen5}{\textbf{Ex5:} \textcolor{orange}{London}'s \textcolor{orange}{mayor Boris Johnson} gives a big thumbs up to \textcolor{orange}{photographers} during the unveiling of the \textcolor{orange}{2012 Olympic rings} on \textcolor{orange}{Tower Bridge}. 
    }} \\

    & &\\
    \noalign{\vskip -7pt}
    \midrule

    \makebox[0.3\textwidth]{\includegraphics[width=0.15\textwidth]{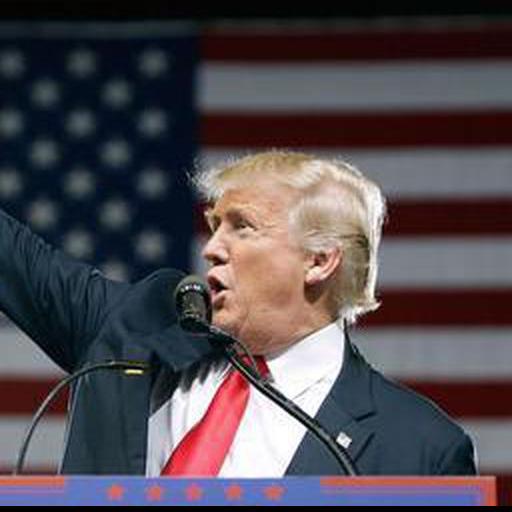}} & \makebox[0.3\textwidth]{\includegraphics[width=0.15\textwidth]{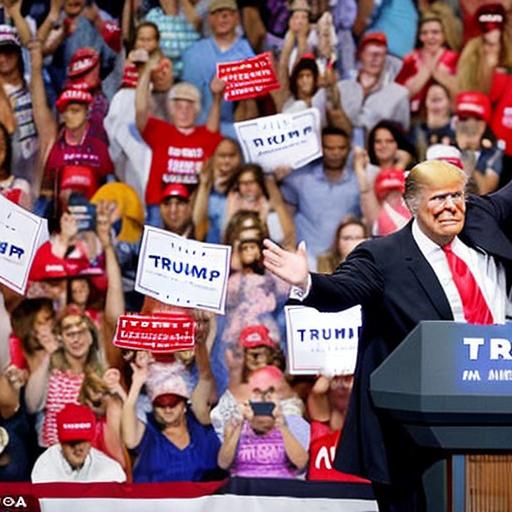}} & \makebox[0.3\textwidth]{\includegraphics[width=0.15\textwidth]{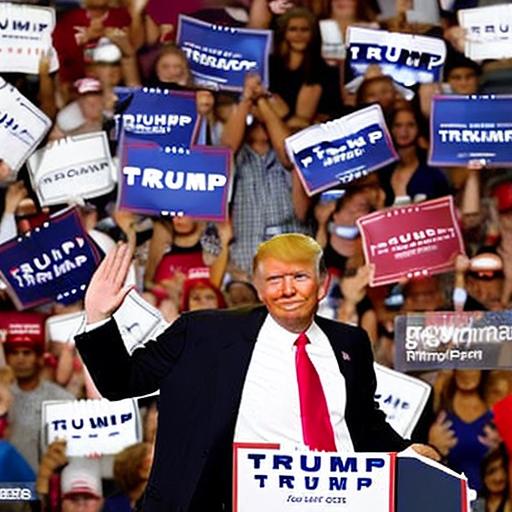}} \\

    \multicolumn{3}{p{0.95\textwidth}}{\centering \hypertarget{fig:suppl-ent-gen6}{\textbf{Ex6:} \textcolor{orange}{Donald Trump} waves to the crowd during a \textcolor{orange}{campaign rally} on \textcolor{orange}{June 18 2016} in \textcolor{orange}{Phoenix.}
    }} \\

    & &\\
    \noalign{\vskip -7pt}
    \midrule

    \makebox[0.3\textwidth]{\includegraphics[width=0.15\textwidth]{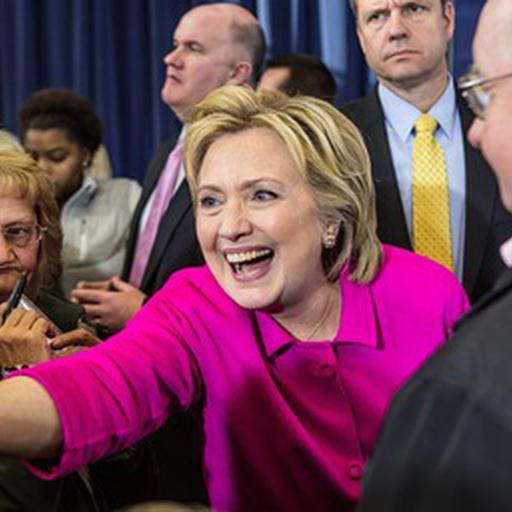}} & \makebox[0.3\textwidth]{\includegraphics[width=0.15\textwidth]{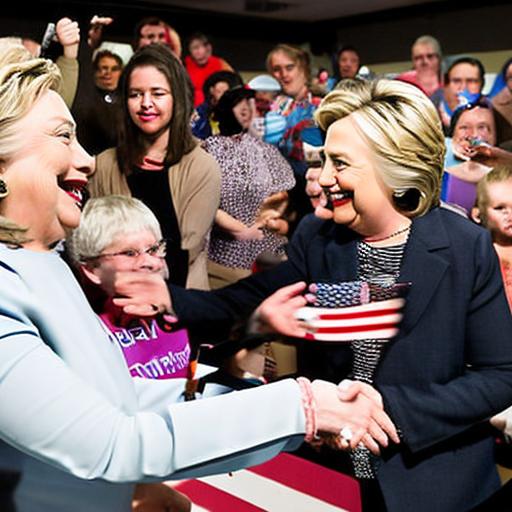}} & \makebox[0.3\textwidth]{\includegraphics[width=0.15\textwidth]{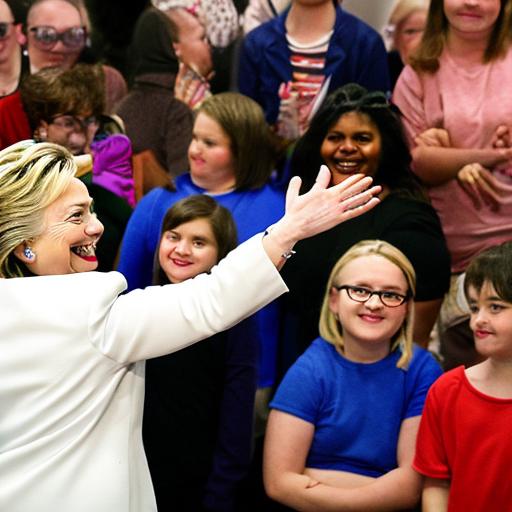}} \\

    \multicolumn{3}{p{0.95\textwidth}}{\centering \hypertarget{fig:suppl-ent-gen7}{\textbf{Ex7:} \textcolor{orange}{Hillary Clinton} greets audience members following a \textcolor{orange}{campaign organizing event} at \textcolor{orange}{Eagle Heights elementary} in \textcolor{orange}{Clinton Iowa}. 
    }} \\

\end{tabular}

\caption{Qualitative comparison of different T2I models on {\n} Entity Set. Words highlighted in \textcolor{orange}{Orange} are used for subject conditioning. Although images generated using {\s} improve on multi-subject representations, the accurate generation of visual features of NEs still serves as a challenge.}
\label{figure:suppl-ent-gen-results}
\end{figure*}

\begin{figure}
      \centering
      \includegraphics[width=0.95\columnwidth]{"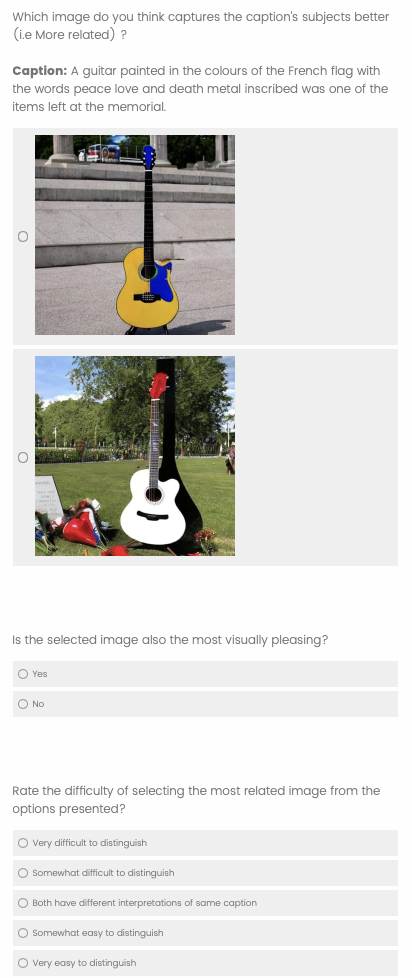"}
      \caption{Survey UI for Generated Image Evaluation Study}
      \label{fig:gen-eval-quest}
\end{figure}

\begin{figure}
      \centering
      \includegraphics[width=0.95\columnwidth]{"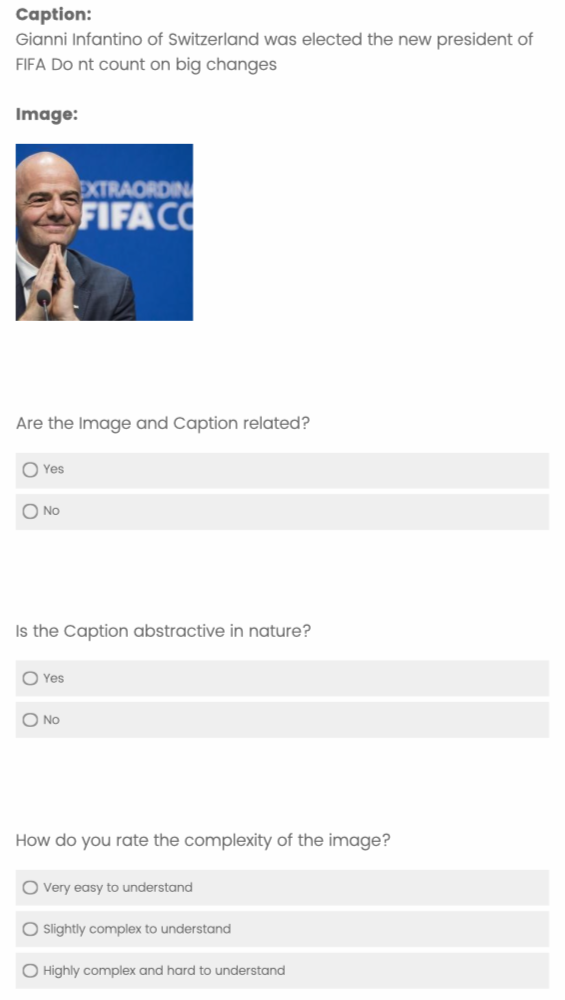"}
      \caption{Survey UI for Data Quality Evaluation Study}
      \label{fig:abstractive-eval-quest}
\end{figure}

\section{Qualitative Evaluation of Generated Samples}
\label{sec:human-evaluation}

For this project, all human evaluation surveys were created on Qualtrics and distributed through Amazon MTurk with our survey UIs provided in Figures \ref{fig:gen-eval-quest}, \ref{fig:abstractive-eval-quest}. All our studies have been conducted with Institutional Review Board (IRB) approval. We do not collect any personally identifiable data from participants in our study. Voluntary consent is obtained from each participant before taking part in all studies. We provide clear instructions for each evaluation task presented to participants with examples and test their understanding using a pre-survey questionnaire. This is done to ensure data quality and improve the consistency of task understanding across participants. Attention Check questions were also incorporated to prevent low-quality submissions from being accepted. The demographic for participants taking part in our survey was limited to people above 18 years of age. 

\subsection{Qualitative Evaluation Setup for {\s}}
\label{sec:qual-eval-setup}

We perform human evaluation of {\s} vs baseline Stable Diffusion on Amazon MTurk to understand perceived variations in subject understanding. From the {\n} Non-Entity test set, we randomly sample and filter 300 captions for our survey. The questions in our survey require participants to pick the image that is most related to the provided caption and also rate the difficulty of choosing between the two images on a 5-point scale. The scale ranges from 1 - "Very easy to distinguish" to 5 - "Very difficult to distinguish". This measure is utilized to understand the rater's confidence in assessing the image-caption pairs. We removed all samples rated as "Very difficult to distinguish" from our analysis to ensure the rating confidence. With each rater labeling a max of 10 samples, we removed all submissions where raters failed our attention checks. Thus, a min of 3 and a max of 5 annotations per sample were present in our final evaluation set. Our analysis shows that raters consistently preferred images generated by {\s} over the baseline model, complementing our quantitative results.

\subsection{Inter-Annotator Agreement}

Additionally, we compute the inter-annotator agreement scores for our MTurk participants to assess the significance of our results. For our human evaluation experiments with SAFE, we use the Krippendorff's $\alpha$ metric which shows an inter-annotator agreement of $\alpha = 0.1216$. This shows a positive correlation between annotators concerning the collected ratings for this task. With the absolute score of $\alpha$ being lower than the average scores reported on other rating tasks, we identify key reasons why this may be the case. In our case, each annotator does not rate every question present in our evaluation samples. So, the unanswered questions by a survey participant are treated as missing values. The high number of missing values when utilizing the typical formulation of this metric is one reason for the lower score observed. Additionally, other studies attempting to assess inter-annotator agreement of T2I generators \cite{Otani2023-no} on complex text-image datasets such as DrawBench \cite{Saharia2022-ui} have reported similarly low scores, indicating the difficulty of this task.

\section{Additional Examples}
\begin{figure*}
\centering
\begin{tabular}{cccc}
    \textbf{Original} & \textbf{SD 2.1 (Base)} & \textbf{SD 2.1 (CR)} & \textbf{SAFE (DFE + GPT-3.5)} \\ \toprule

    \makebox[0.21\textwidth]{\includegraphics[width=0.15\textwidth]{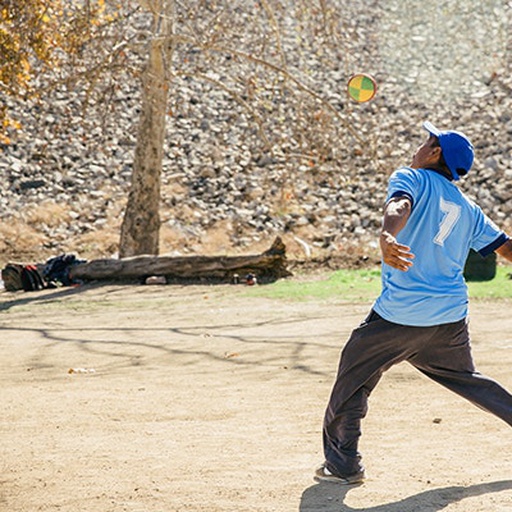}} & \makebox[0.21\textwidth]{\includegraphics[width=0.15\textwidth]{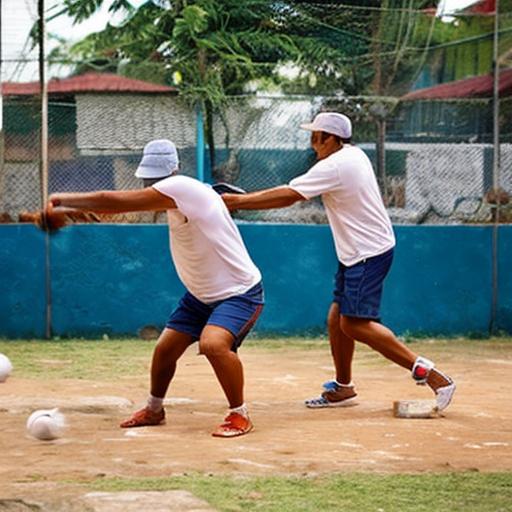}} & \makebox[0.21\textwidth]{\includegraphics[width=0.15\textwidth]{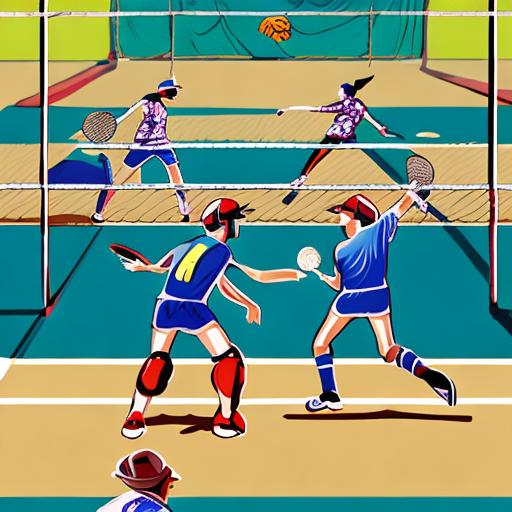}} & \makebox[0.21\textwidth]{\includegraphics[width=0.15\textwidth]{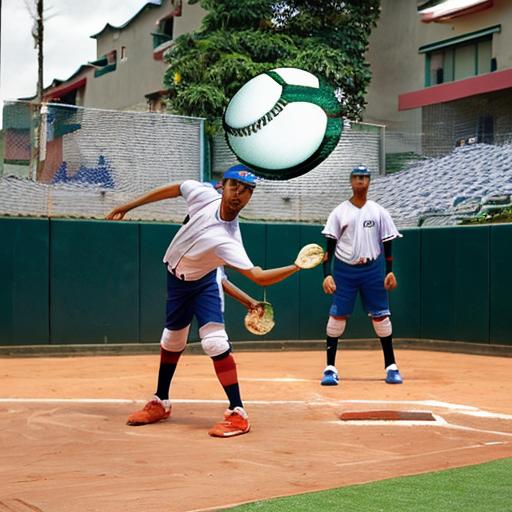}}\\ 

    \multicolumn{4}{p{0.95\textwidth}}{\centering \hypertarget{fig:suppl-nonent-gen8}{\textbf{Ex8:} 
    At first glance, \textcolor{orange}{pelota mixteca} might resemble elements of \textcolor{orange}{baseball}, \textcolor{orange}{volleyball} and \textcolor{orange}{tennis}, but a closer examination reveals a bit more nuance. Each \textcolor{orange}{jugada}, as each individual game is called, involves approximately 10 \textcolor{orange}{players}, and begins when one player initiates a \textcolor{orange}{serve} from the \textcolor{orange}{cement slab}. 
    }} \\

     & & & \\
    \noalign{\vskip -7pt}
    \midrule
    
    \makebox[0.21\textwidth]{\includegraphics[width=0.15\textwidth]{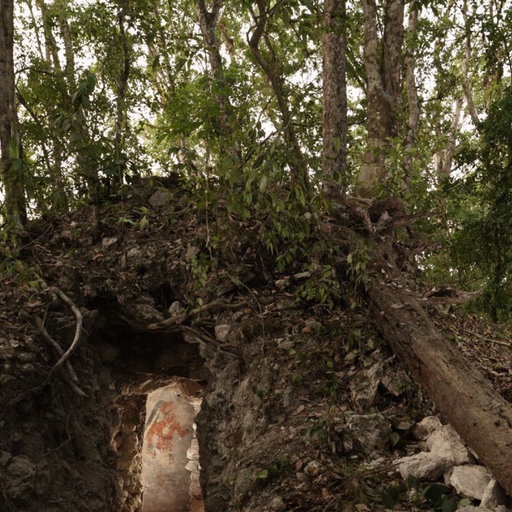}} & \makebox[0.21\textwidth]{\includegraphics[width=0.15\textwidth]{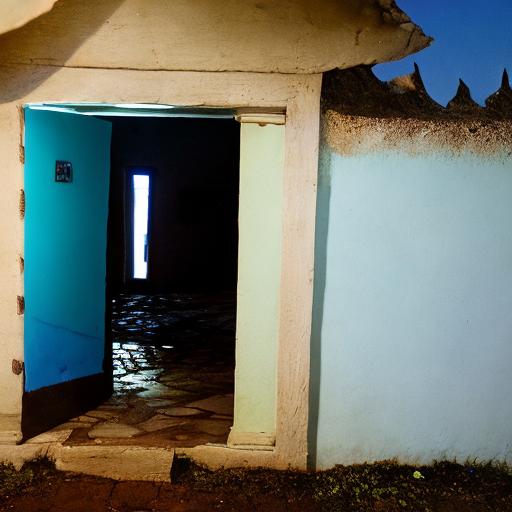}} & \makebox[0.21\textwidth]{\includegraphics[width=0.15\textwidth]{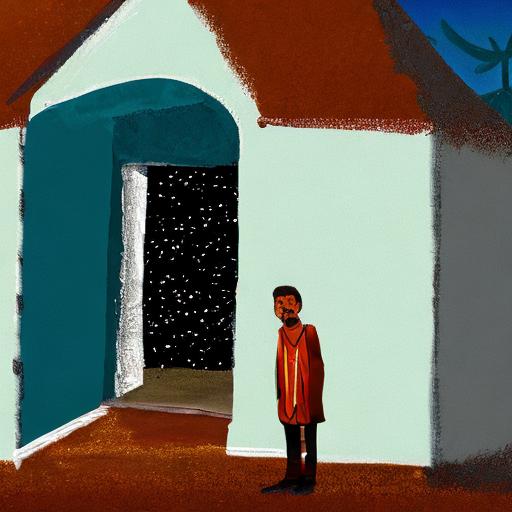}} & \makebox[0.21\textwidth]{\includegraphics[width=0.15\textwidth]{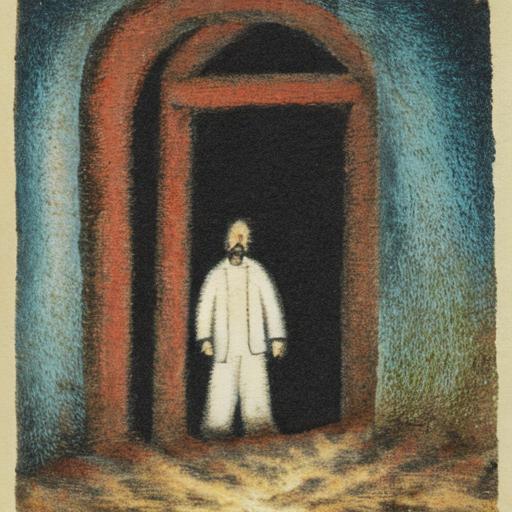}}\\ 

    \multicolumn{4}{p{0.95\textwidth}}{\centering \hypertarget{fig:suppl-nonent-gen9}{\textbf{Ex9:} 
    The \textcolor{orange}{painted figure} of a \textcolor{orange}{man} is \textcolor{orange}{illuminated} through a \textcolor{orange}{doorway} to the \textcolor{orange}{dwelling}.
    }} \\ 

    & & & \\
    \noalign{\vskip -7pt}
    \midrule

    \makebox[0.21\textwidth]{\includegraphics[width=0.15\textwidth]{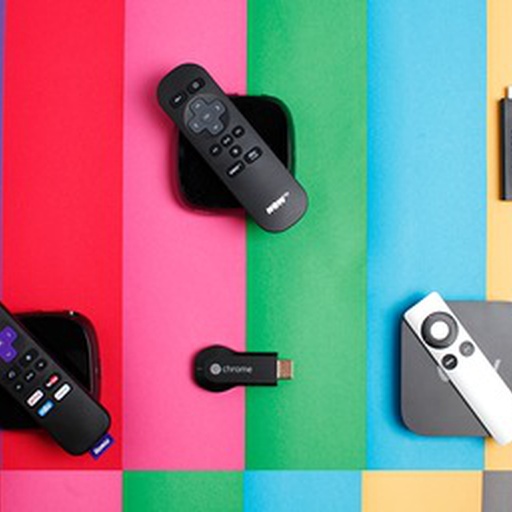}} & \makebox[0.21\textwidth]{\includegraphics[width=0.15\textwidth]{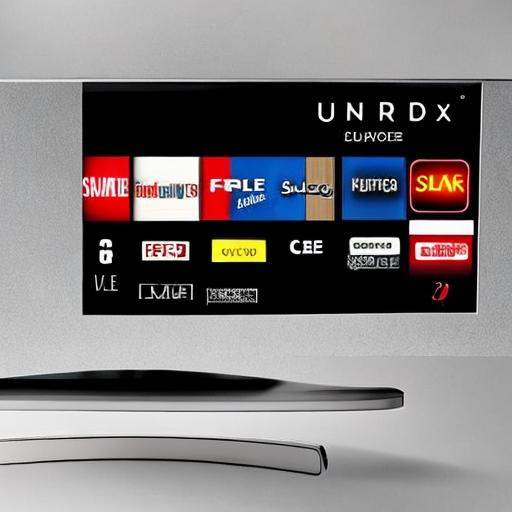}} & \makebox[0.21\textwidth]{\includegraphics[width=0.15\textwidth]{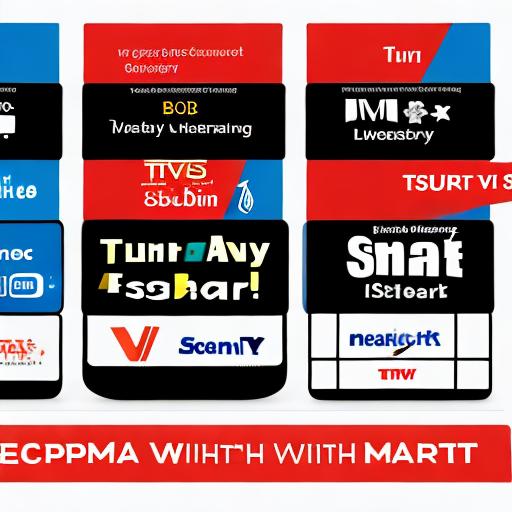}} & \makebox[0.21\textwidth]{\includegraphics[width=0.15\textwidth]{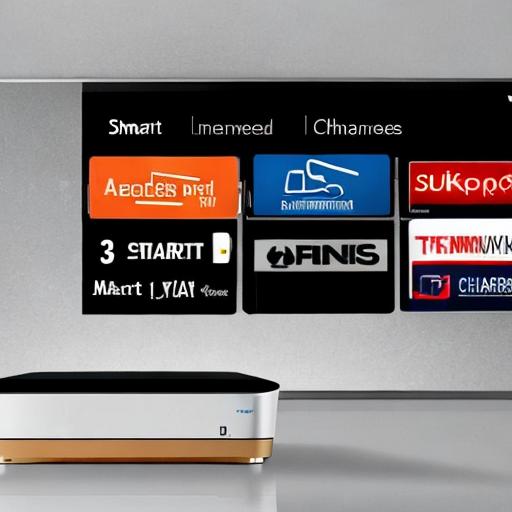}}\\    

    \multicolumn{4}{p{0.95\textwidth}}{\centering \hypertarget{fig:suppl-nonent-gen10}{\textbf{Ex10:} 
    \textcolor{orange}{Mediastreaming boxes} can turn any \textcolor{orange}{TV} \textcolor{orange}{smart} or add \textcolor{orange}{features} and \textcolor{orange}{channels} to others for as little as 15.
    }} \\ 

    \noalign{\vskip -7pt}
    \midrule
    
    \makebox[0.21\textwidth]{\includegraphics[width=0.15\textwidth]{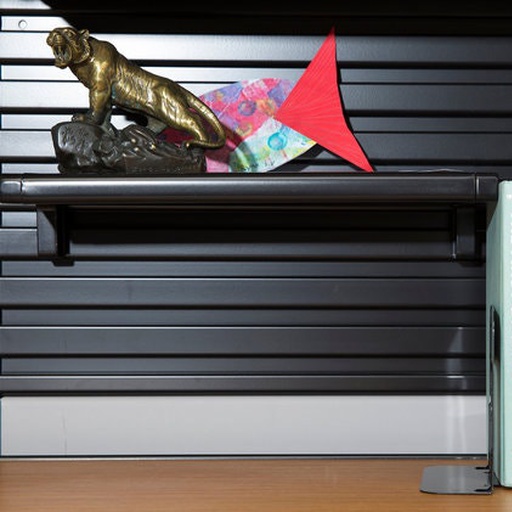}} & \makebox[0.21\textwidth]{\includegraphics[width=0.15\textwidth]{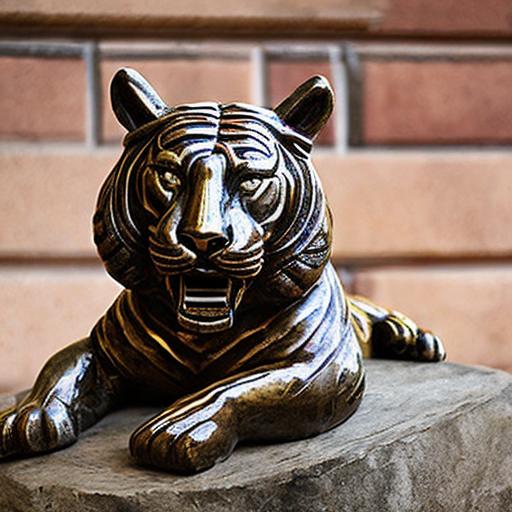}} & \makebox[0.21\textwidth]{\includegraphics[width=0.15\textwidth]{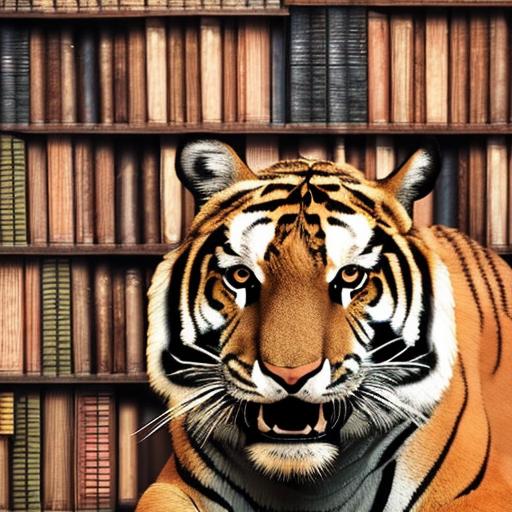}} & \makebox[0.21\textwidth]{\includegraphics[width=0.15\textwidth]{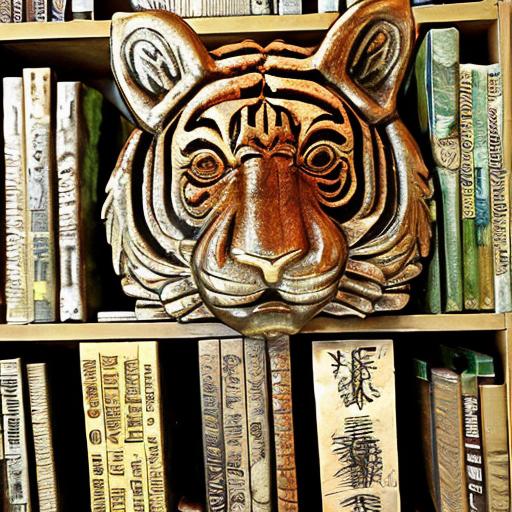}}\\ 
    
    \multicolumn{4}{p{0.95\textwidth}}{\centering \hypertarget{fig:suppl-nonent-gen11}{\textbf{Ex11:} A \textcolor{orange}{bronze tiger} shows \textcolor{orange}{assertiveness} and a \textcolor{orange}{winning spirit}. The \textcolor{orange}{books} are all from \textcolor{orange}{colleagues}.}} \\ 
    & & & \\

    \noalign{\vskip -7pt}
    \midrule
    
    \makebox[0.21\textwidth]{\includegraphics[width=0.15\textwidth]{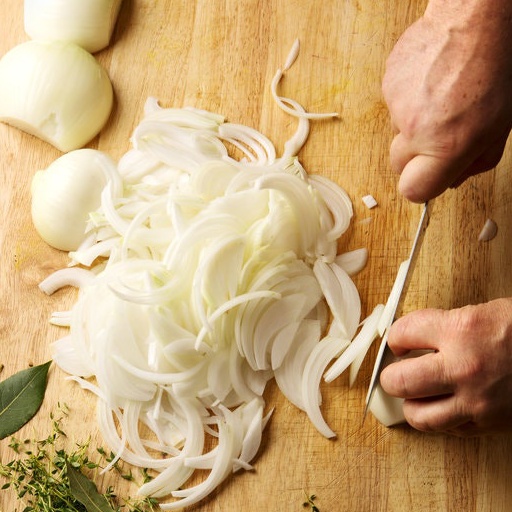}} & \makebox[0.21\textwidth]{\includegraphics[width=0.15\textwidth]{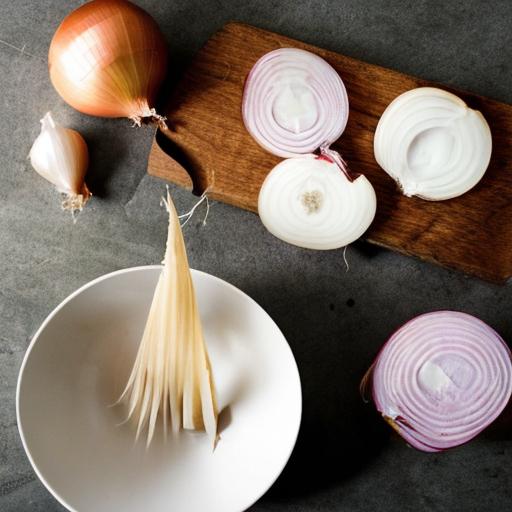}} & \makebox[0.21\textwidth]{\includegraphics[width=0.15\textwidth]{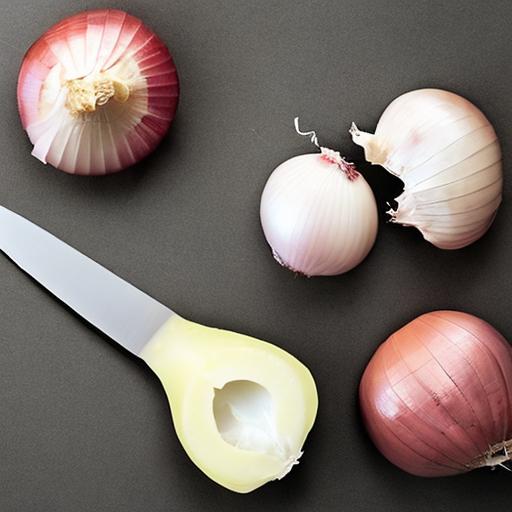}} & \makebox[0.21\textwidth]{\includegraphics[width=0.15\textwidth]{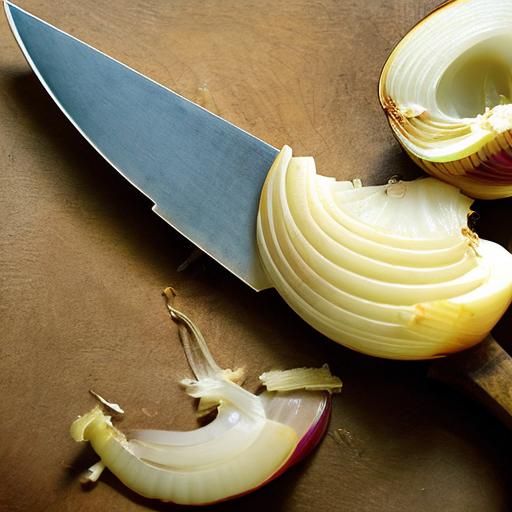}}\\ 
    
    \multicolumn{4}{p{0.95\textwidth}}{\centering \hypertarget{fig:fig:suppl-nonent-gen12}{\textbf{Ex12:}
    A sharp \textcolor{orange}{knife}, one of a \textcolor{orange}{cook}'s essential tools, is used to carefully cut \textcolor{orange}{onions}, which are easier to \textcolor{orange}{brown} (if they’re not bludgeoned) for a \textcolor{orange}{confit}.
        }} \\
        & & & \\

\end{tabular}
\caption{Qualitative comparison of different T2I models on {\n} Non-Entity Subset. Words highlighted in \textcolor{orange}{Orange} are used for subject conditioning}
\label{figure:suppl-nonent-gen-results}
\end{figure*}

We provide additional generated examples using both baseline and SAFE models for reference. Figure \ref{figure:suppl-nonent-gen-results} with examples \hyperlink{fig:suppl-nonent-gen2}{Ex10}, \hyperlink{fig:suppl-nonent-gen3}{Ex11}, \hyperlink{fig:suppl-nonent-gen4}{Ex12} are generated from the test set of {\n} Non-Entity. Similarly \hyperlink{fig:suppl-ent-gen3}{Ex7} and \hyperlink{fig:suppl-ent-gen4}{Ex8} from Figure \ref{figure:suppl-ent-gen-results} are examples from {\n} Entity.

\label{sec:additonal-examples}

\end{document}